\documentclass[lettersize,journal]{IEEEtran}
\usepackage{amsmath,amsfonts}
\usepackage{algorithmic}
\usepackage{algorithm}
\usepackage{array}
\usepackage[caption=false,font=normalsize,labelfont=sf,textfont=sf]{subfig}
\usepackage{textcomp}
\usepackage{stfloats}
\usepackage{url}
\usepackage{verbatim}
\usepackage{graphicx}
\usepackage{xcolor} 
\usepackage{colortbl}
\usepackage{utfsym}
\usepackage{adjustbox}
\usepackage{multirow}
\usepackage{makecell}
\usepackage{booktabs}
\usepackage{titlesec}
\usepackage{hyperref}
\usepackage{refcount}
\usepackage{caption}
\usepackage{threeparttable}
\usepackage[utf8]{inputenc}
\usepackage[numbers,square]{natbib} 
\DeclareMathOperator*{\argmin}{arg\,min}
\begin{document}

\title{Ten Challenging Problems in Federated Foundation Models}

 \author{
      Tao Fan,
      Hanlin Gu,
      Xuemei Cao, 
      Chee Seng Chan,~\IEEEmembership{Senior Member,~IEEE,}
      Qian Chen, 
      Yiqiang Chen,~\IEEEmembership{Senior Member,~IEEE,}
      Yihui Feng,
      Yang Gu,~\IEEEmembership{Member,~IEEE,}
      Jiaxiang Geng,
      Bing Luo,~\IEEEmembership{Senior Member,~IEEE,}
      Shuoling Liu,~\IEEEmembership{Member,~IEEE,}
      Win Kent Ong,
      Chao Ren,~\IEEEmembership{Member,~IEEE,} 
      Jiaqi Shao,
      Chuan Sun,~\IEEEmembership{Member,~IEEE,} 
      Xiaoli Tang,
      Hong Xi Tae,
      Yongxin Tong,~\IEEEmembership{Member,~IEEE,}
      Shuyue Wei,
      Fan Wu,~\IEEEmembership{Member,~IEEE,} 
      Wei Xi,
      Mingcong Xu,
      He Yang,
      Xin Yang,~\IEEEmembership{Member,~IEEE,}
      Jiangpeng Yan,
      Hao Yu,
      Han Yu,~\IEEEmembership{Senior Member,~IEEE,}
      Teng Zhang,
      Yifei Zhang,
      Xiaojin Zhang,
      Zhenzhe Zheng,~\IEEEmembership{Member,~IEEE,} 
      Lixin Fan, ~\IEEEmembership{Member,~IEEE,}
      and Qiang Yang,~\IEEEmembership{Fellow,~IEEE}

\thanks{Apart from the first and corresponding authors, the remaining authors are listed in alphabetical order by their last names.}

\thanks{Tan Fan, Hanlin Gu are with the WeBank, Shenzhen, China (\textit{Co-First Author}, e-mail: tfanac@cse.ust.hk, allengu@webank.com).}

\thanks{Chee Seng Chan, Win Kent Ong, Hong Xi Tae are with the Universiti Malaya, Malaysia (e-mail: \{cs.chan, winkent.ong\}@um.edu.my, taehongxi55@gmail.com).}

\thanks{Yiqiang Chen, Yang Gu, Qian Chen, Teng Zhang are with the Institute of Computing Technology, Chinese Academy of Sciences, China (e-mail:  \{yqchen, guyang, chenqian20b, zhangteng19s\}@ict.ac.cn).}

\thanks{Bing Luo, Jiaxiang Geng, Jiaqi Shao are with the Duke Kunshan University, Kunshan, China (e-mail: \{bl291, jg645,  js1139\}@duke.edu).}

\thanks{Shuoling Liu, Jiangpeng Yan are with the Innovation Reseach Center, EFunds, China (e-mail: \{liushuoling, yanjiangpeng\}@efunds.com.cn).}

\thanks{Chao Ren is with the KTH Royal Institute of Technology, Sweden (e-mail: renc0003@e.ntu.edu.sg).}

\thanks{Yongxin Tong, Shuyue Wei are with the Beihang University, China (e-mail: \{yxtong, weishuyue\}@buaa.edu.cn).}

\thanks{Fan Wu, Zhenzhe Zheng are with the Shanghai Jiao Tong University, China (e-mail: fwu@cs.sjtu.edu.cn, zhengzhenzhe@sjtu.edu.cn).}

\thanks{Wei Xi, He Yang are with the Xi'an Jiaotong University, China (e-mail: xiwei@xjtu.edu.cn, sleepingcat@stu.xjtu.edu.cn).}

\thanks{Xin Yang, Xuemei Cao, Yihui Feng, Hao Yu are with the Southwestern University of Finance and Economic, China (e-mail: yangxin@swufe.edu.cn, caoxuemei.qpz@gmail.com, yihuifeng@foxmail.com, yuhao2033@163.com).}

\thanks{Han Yu, Chuan Sun, Xiaoli Tang, Yifei Zhang are with the Nanyang Technological University, Singapore (e-mail: \{han.yu, chuan.sun,  yifei.zhang\}@ntu.edu.sg, xiaoli001@e.ntu.edu.sg).}

\thanks{ Xiaojin Zhang, Mingcong Xu are with the Huazhong University of Science and Technology, China (e-mail: \{xiaojinzhang, xumingcong\}@hust.edu.cn).}

\thanks{Lixin Fan is the Principal Scientist of Artificial Intelligence at
WeBank, Shenzhen, China (\textit{Co-Corresponding Author}, e-mail: lixinfan@webank.com).} 

\thanks{Qiang Yang is Professor Emeritus at the Department of Computer Science and Engineering,
Hong Kong University of Science and Technology, Hong Kong, and the Chief AI Officer of WeBank, Shenzhen, China (\textit{Co-Corresponding Author}, e-mail: qyang@cse.ust.hk).} 
}


\maketitle
\begin{abstract}
Federated Foundation Models (FedFMs) represent a distributed learning paradigm that fuses general competences of foundation models as well as privacy-preserving capabilities of federated learning. This combination allows the large foundation models and the small local domain models at the remote clients to learn from each other in a teacher-student learning setting. This paper provides a comprehensive summary of the ten challenging problems inherent in FedFMs, encompassing foundational  theory, utilization of private data, continual learning, unlearning, Non-IID and graph data, bidirectional knowledge transfer, incentive mechanism design, game mechanism design, model watermarking, and efficiency.  
The ten  challenging problems manifest in five pivotal aspects: ``Foundational Theory,"  which aims to establish a coherent and unifying theoretical framework for FedFMs. ``Data," addressing the difficulties in leveraging domain-specific knowledge from private data while maintaining privacy; ``Heterogeneity," examining variations in data, model, and computational resources across clients; ``Security and Privacy," focusing on defenses against malicious attacks and model theft; and ``Efficiency," highlighting the need for improvements in training, communication, and parameter efficiency. 
For each problem, we offer a clear mathematical definition on the objective function, analyze existing methods, and discuss the key challenges and potential solutions. This in-depth exploration aims to advance the theoretical foundations of FedFMs, guide practical implementations, and inspire future research to overcome these obstacles, thereby enabling the robust, efficient, and privacy-preserving FedFMs in various real-world applications.
\end{abstract}


\IEEEpeerreviewmaketitle

\section{introduction}

\subsection{Motivation}

Foundation Models (FMs) \cite{zhou2024comprehensive} have emerged as a groundbreaking force in the realm of artificial intelligence. Renowned FMs, including GPT-4 and LLaMa, have exhibited an extraordinary ability to understand context and nuances, enabling them to skillfully handle a wide range of tasks across diverse fields such as natural language processing (NLP) and computer vision (CV).
On the other hand, Domain Models (DMs), often deployed  remotely on edge devices, are trained locally using private data that cannot be shared with the FMs. However, DMs also have limitations stemming from their restricted generalization capacities.
The dilemma  raises a question: \textit{How can we effectively harness the power of FMs for domain-specific knowledge while simultaneously ensure that DMs possess adequate generalization capabilities and adhere to privacy protection requirements?}

\begin{figure}
    \centering
    \includegraphics[width=0.9\linewidth]{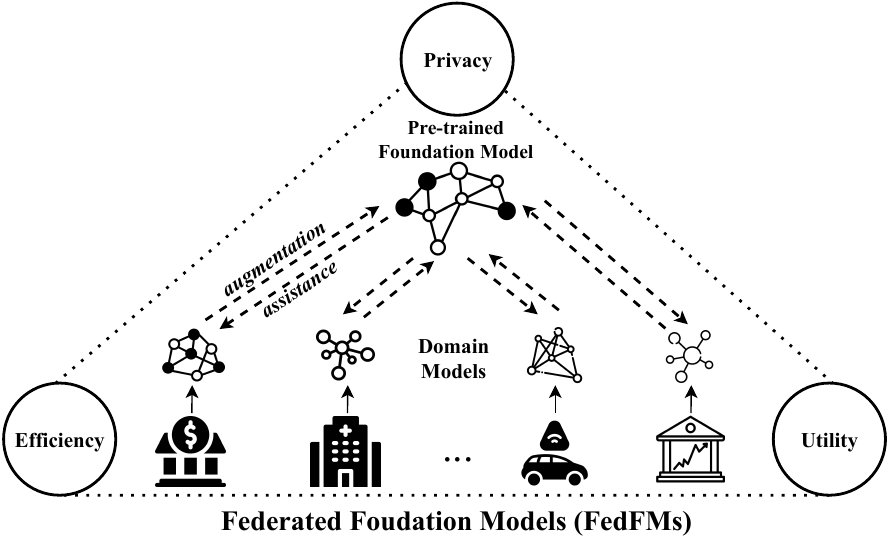}
    \caption{An illustration of Federated Foundation Models (FedFMs). 
    On one hand, Domain Models (DMs) can augment the domain-specific knowledge of Foundation Models (FMs) through FedFMs. On the other hand, FMs can assist in enhancing the generalization capabilities of DMs on the edges in a distributed setting.}
    \label{fig:fedfms}
    \vspace{-5mm}
\end{figure}

One promising solution to integrate capabilities of FMs and DMs is through Federated Foundation Models (FedFMs)~\cite{fan2023fate,kang2023grounding,ren2024advances}.  This paper provides a comprehensive summary of definitions and challenges in this new machine learning paradigm. Specifically, as depicted in Fig. \ref{fig:fedfms}, FedFMs is defined as a distributed learning framework which includes at least one pre-trained FM and a multitude of DMs. 
Federated learning (FL) \cite{yang2019federated} methods are adopted in FedFMs to ensure that data privacy are well-respected during the training of FedFMs, especially, for healthcare, finance and IoTs  applications.

\captionsetup[table]{labelformat=simple, labelsep=newline, textfont=sc, justification=centering}

\begin{table*}[ht]
		\caption{Ten Challenging Problems in FedFMs} 
		\label{tab:framework}
		\resizebox{\textwidth}{!}{
			\begin{threeparttable}
				\begin{tabular}{|l|l|l|l|}
					\specialrule{0em}{2pt}{2pt} 
					\hline \hline
					\textbf{\qquad \qquad \qquad \qquad Problems}                                                                 & \multicolumn{1}{c|}{ \textbf{Existing Methods}}                                                                  & \multicolumn{1}{c|}{ \textbf{Challenges}} & \multicolumn{1}{c|}{ \textbf{Potential Solutions}}
					      \\ \hline
					\begin{tabular}[l]{@{}c@{}} Problem 1: How to establish a foundational theory for FedFMs?\end{tabular}               
					& \begin{tabular}[c]{@{}l@{}}
						$\bullet$ Multi-objective Trade-off Methods \cite{zhang2024no,dong2023tunable,hou2023privately,kuang2024federatedscope,wu2024fedbiot}\\
						$\bullet$ Open- and Closed-Source FMs Trade-off Methods \cite{zhang2023towards, wang2024flora,guo2023pfedprompt, yang2023efficient}\end{tabular}                                                       & \begin{tabular}[c]{@{}l@{}}
						$\bullet$ Multi-Objective Optimization in FedFMs\\ 
						$\bullet$ Fairness in FedFMs \\ 
						$\bullet$ Black-Box and White-Box \\Aggregation in FedFMs 
					\end{tabular} 

                     & \begin{tabular}[c]{@{}l@{}}
                          $\bullet$ Establishing Theoretical Bounds\\
                          $\bullet$ Balancing Cross-Layer \\~Privacy-Utility-Efficiency \\ 
					   $\bullet$ Adaptive Strategies \\ 
						 $\bullet$ Hybrid Aggregation Techniques
					\end{tabular} 
					
					\\ \hline
					\begin{tabular}[l]{@{}l@{}} Problem 2: How to utilize private data in FedFMs?\end{tabular}             & \begin{tabular}[c]{@{}l@{}}
						$\bullet$ Data-Centric Methods  \cite{6-data-dpsda,qin2023federated,6-data-fedkim,6-data-fedcampus}\\
						$\bullet$ Model-Centric Methods \cite{zhang2024fedpit,ye2024openfedllm,JCST-2308-13639,zhangenhancing}\end{tabular}                        
					& \begin{tabular}[l]{@{}l@{}}
						$\bullet$ Data Quality Impact\\ 
						$\bullet$ Scale and Efficiency Trade-offs  \\ 
						$\bullet$ Computational Resource Constraints \end{tabular}

                    & \begin{tabular}[l]{@{}l@{}}
						$\bullet$ Balancing Data Efficiency \\and Computational Cost \\
						$\bullet$ Enhancing Private Data Processing  \\ 
						$\bullet$ Refining Data Contribution Evaluation 
                        \end{tabular}
					
					\\ \hline
					\begin{tabular}[l]{@{}l@{}}Problem 3: How to design continual learning in FedFMs?\end{tabular}       
					& \begin{tabular}[c]{@{}l@{}}
						$\bullet$ Replay-based Methods \cite{li2024towards,li2024sr,yu2024overcoming,liang2025diffusion}  \\
						$\bullet$ Methods based on Regularization and Decomposition \cite{yoon2021federated,dong2022federated}\\
                        $\bullet$ Distillation-based Methods \cite{ma2022continual,wei2022knowledge}\\
						$\bullet$ Prompted-based Methods \cite{piaofederated,yu2024personalized} 
					\end{tabular}                          
					
					& \begin{tabular}[c]{@{}l@{}}
						$\bullet$ Spatial-Temporal Data Heterogeneity \\ 
                        $\bullet$ Knowledge Transfer and Error Correction \\
                        $\bullet$ Knowledge Conflict \\
                        \end{tabular} 

                    & \begin{tabular}[c]{@{}l@{}}
                        $\bullet$ External Knowledge Base\\
                        $\bullet$ Selective Knowledge Fusion\\
                        \end{tabular} 
					                                           
					\\ \hline
					\begin{tabular}[l]{@{}l@{}} Problem 4: How to design unlearning in FedFMs?\end{tabular}                                                             
					& \begin{tabular}[c]{@{}l@{}}
						$\bullet$ Unlearning Targets \cite{wang2022federated, shah2023unlearning, gu2024few,gu2024ferrari,Gu2024Unlearning}\\
						$\bullet$ Unlearning Executors \cite{liu2022right,su2023asynchronous, fraboni2024sifu,halimi2022federated,wu2022federated,gao2024verifi}\\
                        $\bullet$ Unlearning Verification \cite{cao2023fedrecover,zhang2023fedrecovery}\\
						$\bullet$ Unlearning Principles  \cite{shao2024federated}\end{tabular}                                     
					
					& \begin{tabular}[c]{@{}l@{}}
						$\bullet$ High Model Complexity\\
						$\bullet$ Knowledge Coupling \\
                            $\bullet$ Cross-Client Consistency
                        \end{tabular} 

                    & \begin{tabular}[c]{@{}l@{}}
						$\bullet$ Modular Optimization Unlearning\\
						$\bullet$ Disentangled Knowledge Decoupling \\
                            $\bullet$ Federated Cross-Client Coordination
                        \end{tabular} 
                        
				\\ \hline
					\begin{tabular}[l]{@{}l@{}} Problem 5: How to address NON-IID issues \\ and graph data in FedFMs?\end{tabular}                   
                    & \begin{tabular}[c]{@{}l@{}}
						$\bullet$ Distribution Adaptation \cite{park2024fedbaffederatedlearningaggregation,imteaj2024tripleplayenhancingfederatedlearning,6-data-feddpa}\\ 
						$\bullet$ Federated Graph of Models \cite{li2024fedgtatopologyawareaveragingfederated,ma2024beyond,huang2022accelerating}\\ 
						$\bullet$ Optimization in Graph of Models \cite{chen2024privfusion, chen2024model}\end{tabular}                                          
					
					& \begin{tabular}[c]{@{}l@{}}
						$\bullet$ Convergence Issues from Heterogeneity\\ 
						$\bullet$ Complexity of Network Topology \\
                            $\bullet$ Sustainable Optimization
                        \end{tabular}

                    & \begin{tabular}[c]{@{}l@{}}
						$\bullet$ Adaptive Optimization\\ 
						$\bullet$ Dynamic and Topology-aware \\ Aggregation \\
                            $\bullet$ Multi-objective Optimization
                            \end{tabular}
                        
					                               \\ \hline
					\begin{tabular}[l]{@{}l@{}}  Problem 6: How to achieve the bidirectional knowledge transfer \\ ~between FMs and DMs in FedFMs? \end{tabular}              
					& \begin{tabular}[l]{@{}l@{}}
						$\bullet$ Data-Level Knowledge Transfer \cite{hsieh2023distilling,li2022explanations,jiang2023lion,fan2024pdss,li2024federated}\\ 
						$\bullet$ Representation-Level Knowledge Transfer \cite{shen2023split,he2019model,gu2024minillm,yu2023multimodal,fan-etal-2025-fedmkt}\\
						$\bullet$ Model-Level Knowledge Transfer \cite{zhang2023gpt,xiao2023offsite,fan2024fedcollm,deng2023mutual}
					\end{tabular}                                                        
                    & \begin{tabular}[l]{@{}l@{}}
						$\bullet$  Data Heterogeneity\\ 
						$\bullet$  Representation Heterogeneity\\ 
                            $\bullet$  Model Heterogeneity\\ 
						$\bullet$  Privacy\\
                        \end{tabular} 
                        
					& \begin{tabular}[l]{@{}l@{}}
						$\bullet$ Unified Model Architectures\\ 
						$\bullet$ Adaptive Knowledge Transfer\\ 
                         $\bullet$ Synthetic Data\\
                        $\bullet$ Advanced Privacy Techniques
                        \end{tabular} 
					                                                                     \\ \hline
					\begin{tabular}[l]{@{}l@{}}  Problem 7: How to design incentive mechanisms\\ through contribution evaluation in FedFMs? \end{tabular}             
					& \begin{tabular}[c]{@{}l@{}}
						$\bullet$ Individual based Evaluation Schemes~\cite{Individual}\\
                        $\bullet$ Leave-One-Out based Evaluation Schemes~\cite{LOO1999} \\
						$\bullet$ Interaction based Evaluation Schemes~\cite{RN9, RN33} \\ 
						$\bullet$ Shapley-Value based Contribution Evaluation~\cite{song2019profit, wei2020efficient,RN21} \\
                        $\bullet$ Least Core based Evaluation Schemes~\cite{Yan2021IfYL} 
                        \end{tabular} 
					& \begin{tabular}[c]{@{}l@{}}
						$\bullet$ Scalability and Computational Overhead\\ 
						$\bullet$ Data-Task Contribution Mapping \end{tabular} 

                    & \begin{tabular}[c]{@{}l@{}}
						$\bullet$ Learned Contribution Evaluation\\ 
						$\bullet$ Crowd-sourced Data Contribution Map \end{tabular} 
                        
                        \\ \hline 
					\begin{tabular}[l]{@{}l@{}} Problem 8: How to design game mechanisms in FedFMs? \end{tabular} 
					
					& \begin{tabular}[c]{@{}l@{}}
						$\bullet$ Stackelberg Game-based Methods \cite{sarikaya2019motivating,pandey2020crowdsourcing,khan2020federated,luo2023incentive,hu2020trading,lee2020market}\\ $\bullet$ Yardstick Competition-based Schemes \cite{sarikaya2020regulating}\\
                        $\bullet$ Shapley-Value Based Schemes \cite{qu2020privacy,song2019profit}\\
                        $\bullet$ Other Incentive Mechanisms \cite{tang2023utility,tang2023competitive,tang2024bias,tang2023multi,tang2024cost,tang2024intelligent,tang2024stakeholder}
                        \end{tabular}                      
					& \begin{tabular}[c]{@{}l@{}}
						$\bullet$ Privacy Cost Quantification \\
						$\bullet$ Information Asymmetry \\
						$\bullet$ Multi-objective Tradeoffs \end{tabular}     

                    & \begin{tabular}[c]{@{}l@{}}
						$\bullet$ Adaptive Game-theoretic Models\\
						$\bullet$ Resource-aware Privacy Mechanisms\\
						$\bullet$ Incentive-compatible Mechanisms\\
                            $\bullet$ AI-driven Anomaly Detection\end{tabular}

					\\  \hline
					\begin{tabular}[l]{@{}l@{}} Problem 9: How to design model watermarking in FedFMs?  \end{tabular}    
					& \begin{tabular}[c]{@{}l@{}}
						$\bullet$ White-Box Watermarking \cite{li2022fedipr,yang2023fedsov,shao2024fedtrackerfurnishingownershipverification,liang2023fedcip}\\ 
						$\bullet$ Black-Box Watermarking \cite{atli2021wafflewatermarkingfederatedlearning, 9658998, yang2023watermarkingsecurefederatedlearning, electronics13214306}\end{tabular}                                                                             & \begin{tabular}[c]{@{}l@{}}
						$\bullet$ Large-Scale Watermarking \\ 
						$\bullet$ Scalability with Massive FMs\\ 
						$\bullet$ Multi-Modalities  \\
                        $\bullet$ Dynamic Model Composition
                        \end{tabular}
                      &
                    \begin{tabular}[c]{@{}l@{}}
						$\bullet$ Lightweight Watermark Embedding\\ and Federated Pruning\\ 
						$\bullet$ Resource-efficient Watermarking\\ 
						$\bullet$ Modality-agnostic Watermarking \\
                        $\bullet$ Adaptive Watermarks \\with Self-reinforcing Signatures
                        \end{tabular}    
                        
					\\  \hline
					\begin{tabular}[l]{@{}c@{}} Problem 10: How to to improve the efficiency in FedFMs? \end{tabular}       
					& \begin{tabular}[c]{@{}l@{}}
                        $\bullet$ Data Selection Strategies and Prompt Compression \cite{ye2024openfedllm,qin2024federated}\\
                        $\bullet$ Model Compression \cite{ren2023two,fan2024data,li2019fedmd}\\
                        $\bullet$ Resource-aware Scheduling and Framework Optimization \cite{hou2020dynabert,niu2024smartmem}\\
                        $\bullet$ Communication Parameter Reduction \cite{wang2021resource, hu2021mhat}\end{tabular}    
                    
                        & \begin{tabular}[c]{@{}l@{}}
						$\bullet$ Communication Bottlenecks\\ 
                        $\bullet$ Efficiency-Performance-Privacy Trade-off\\ 
						$\bullet$ Task Redundancy \end{tabular}

                        & \begin{tabular}[c]{@{}l@{}}
						$\bullet$ Compression\\ 
                        $\bullet$ Hybrid Privacy-Preserving Techniques\\ 
						$\bullet$ Disentangling Tasks\end{tabular}

					\\  \hline 
				\end{tabular} 
				
		\end{threeparttable} }
	\end{table*}

\subsection{Ten Challenging Problems in FedFMs}

Despite their promising applications, FedFMs face significant challenges for widespread adoption. 
In this paper, we delve into the intrinsic challenges of FedFMs from five key aspects: “\textbf{Foundational Theory}”, “\textbf{Data}”, “\textbf{Heterogeneity}”, “\textbf{Security and Privacy}” and “\textbf{Efficiency}”.

\textbf{Foundational Theory} addresses the optimization objectives at the highest level in FedFMs. The theory must address issues such as how to harmonize privacy, utility, and efficiency within the framework of FedFMs \cite{zhang2022no}, guiding algorithm design (\textit{Problem 1}). 

\textbf{Data}, challenges arise from data distribution across organizations or individuals. FedFMs must adapt to unique private data characteristics, balancing data utility and privacy risks \cite{kang2023grounding} (\textit{Problem 2}). Data also varies over time; thus, FedFMs must adapt to new task requirements without forgetting previously learned knowledge \cite{yang2024federatedcl} (\textit{Problem 3}). Additionally, FedFMs need mechanisms for “unlearning” when participants exit \cite{liu2024survey} (\textit{Problem 4}).

\textbf{Heterogeneity} in FedFMs includes data, model, and computational aspects. Non-IID data variations across clients challenge effective generalization and pattern capture, necessitating advanced methodologies \cite{park2024fedbaffederatedlearningaggregation} (\textit{Problem 5}). Additionally, collaboration between large FMs and small domain models introduces bidirectional knowledge transfer challenges \cite{fan-etal-2025-fedmkt} (\textit{Problem 6}). Lastly, evaluating and fairly rewarding participants' contributions remain critical \cite{wei2020efficient} (\textit{Problem 7}).

\textbf{Security and Privacy} are vital for FedFMs' practical application. FedFMs face two primary security threats: privacy attacks and local data/model theft. Addressing these requires both defensive technical measures and game-theoretic mechanisms to deter attackers \cite{zhang2024game} (\textit{Problem 8}). To combat model theft, leveraging model watermarking for full life-cycle tracking, auditing, and detection is crucial  \cite{li2022fedipr} (\textit{Problem 9}).

\textbf{Efficiency} refers to optimized training and inference times and computational resources. There are still significant improvements needed in three key areas to ensure efficiency: training efficiency, communication efficiency, and parameter efficiency 
 \cite{mcmahan2017communication,wang2021resource, hu2021mhat} (\textit{Problem 10}).

This paper critically assesses recent advancements in FedFMs from the aforementioned five aspects and identifies the \textbf{\textit{ten challenging problems}}, summarized in TABLE \ref{tab:framework}.

\subsection{Related Works}

The research field of FedFMs is in its early stages, with only two position papers \cite{chen2023federated,li2024position} and six review papers \cite{zhuang2023foundation,yu2023federated,kang2023grounding,woisetschlager2024survey,li2024synergizing,ren2024advances} available. Table \ref{tab:existingreview} compares these papers to our survey, indicating coverage of key FedFMs aspects (\usym{1F5F8} mentioned, \usym{2717} not discussed). While existing surveys provide some summary of FedFMs, they are often concise and lack comprehensiveness. To our knowledge, this survey presents the most exhaustive treatment of the FedFMs topic to date.

\subsection{Organization of This Paper}

The remainder of this paper is organized as follows.
Sections II-XI address \textit{Problem 1-10}, each dedicated to a single problem. For each problem, we provide a concise definition, review existing methods, and explore key challenges and potential solutions.
Section \ref{sec:summay} summarizes key insights from the 10 challenging problems in FedFMs, providing readers with a clear understanding of practical implications and the broader context of the findings. Notations used in this paper are shown in Table \ref{tab:notation}.

\begin{table*}[htbp]
\setlength\tabcolsep{0.6pt}
\caption{Comparison of Existing Reviews and Position Papers with This Survey Paper}
\label{tab:existingreview}
\centering
\resizebox{\textwidth}{!}{
\begin{tabular}{c|c|cccccccccc}
\toprule
{\textsc{Years}} & {\textsc{References}} & 
{\makecell[c]{\textsc{Problem} \\ \textsc{1}} }& 
{\makecell[c]{\textsc{Problem} \\ \textsc{2}}}& 
{\makecell[c]{\textsc{Problem} \\ \textsc{3}}}&
{\makecell[c]{\textsc{Problem} \\ \textsc{4}}}& 
{\makecell[c]{\textsc{Problem} \\ \textsc{5}}}&
{\makecell[c]{\textsc{Problem} \\ \textsc{6}}}& 
{\makecell[c]{\textsc{Problem} \\ \textsc{7}}}&
{\makecell[c]{\textsc{Problem} \\ \textsc{8}}}&
{\makecell[c]{\textsc{Problem} \\ \textsc{9}}}&
{\makecell[c]{\textsc{Problem} \\ \textsc{10}}}
\\

\midrule
\multicolumn{1}{c|}{\multirow{4}{*}{\begin{tabular}[c]{@{}l@{}}2023\end{tabular}}} &
Chen \textit{et al}. \cite{chen2023federated} & \usym{2717}   & \usym{1F5F8} & \usym{2717} & \usym{2717}  & \usym{1F5F8} & \usym{2717} & \usym{1F5F8} & \usym{2717} & \usym{1F5F8} & \usym{1F5F8} \\ 
\multicolumn{1}{c|}{} &
Zhuang \textit{et al}. \cite{zhuang2023foundation} & \usym{2717}   & \usym{1F5F8} & \usym{2717} & \usym{2717}  & \usym{1F5F8} & \usym{2717} & \usym{1F5F8} & \usym{2717} & \usym{1F5F8} & \usym{1F5F8} \\ 
\multicolumn{1}{c|}{} &
Yu \textit{et al}. \cite{yu2023federated} & \usym{2717}   & \usym{1F5F8} & \usym{1F5F8} & \usym{2717} & \usym{2717} & \usym{2717} & \usym{2717} & \usym{2717} & \usym{2717} & \usym{1F5F8} \\ 
\multicolumn{1}{c|}{} &
Kang \textit{et al}. \cite{kang2023grounding} & \usym{2717}  & \usym{1F5F8} & \usym{2717}  & \usym{2717}  & \usym{2717} & \usym{1F5F8}   & \usym{2717} & \usym{2717} & \usym{2717} & \usym{1F5F8}   \\ \midrule

\multicolumn{1}{c|}{\multirow{4}{*}{\begin{tabular}[c]{@{}l@{}}2024\end{tabular}}} &
Herbert \textit{et al}. \cite{woisetschlager2024survey} &  \usym{2717}  & \usym{1F5F8}   & \usym{2717}  & \usym{2717}  & \usym{2717}  & \usym{2717}   & \usym{2717}  &  \usym{2717} & \usym{2717} & \usym{1F5F8}   \\ 
\multicolumn{1}{c|}{} &
Li \textit{et al}. \cite{li2024position} & \usym{2717} & \usym{1F5F8} & \usym{2717} & \usym{2717}  & \usym{1F5F8}   & \usym{2717} & \usym{2717}  & \usym{2717} & \usym{2717} & \usym{2717}\\ 
\multicolumn{1}{c|}{} &
Li \textit{et al}. \cite{li2024synergizing} & \usym{2717}  & \usym{1F5F8} & \usym{2717}  & \usym{2717}  & \usym{2717} & \usym{2717}  & \usym{2717}  & \usym{2717}  & \usym{1F5F8} & \usym{1F5F8} \\ 
\multicolumn{1}{c|}{} &
Ren \textit{et al}. \cite{ren2024advances} & \usym{2717}  & \usym{1F5F8} & \usym{2717}  & \usym{2717}  & \usym{1F5F8}  & \usym{2717} & \usym{1F5F8} & \usym{2717} & \usym{1F5F8}  & \usym{1F5F8}  \\ 

\midrule

\multicolumn{2}{c|}{\textbf{This paper}} & 
 \usym{1F5F8}   & \usym{1F5F8} & \usym{1F5F8} & \usym{1F5F8}  & \usym{1F5F8} & \usym{1F5F8} & \usym{1F5F8} & \usym{1F5F8} & \usym{1F5F8} & \usym{1F5F8}\\ 

\bottomrule 
\end{tabular}
}
\end{table*}

\begin{table}[htbp]
\normalsize
\caption{Table of Notation}
    \begin{center}
        \begin{tabular}{lccc}
            \hline
            Notation & Meaning \\ 
            \hline 
            $K$ & number of clients \\
            $D_k$ & private dataset of client $k$ \\
                        $D_p$ & public dataset \\
            $S_k$ & data distribution of $D_k$ \\
            $\ell_u$  & utility loss \\
            $\ell_c$  & contribution loss \\ 
            $\ell_{m}$  & watermark loss \\
            $\ell_p$ &privacy loss\\
            $\ell_e$ & efficiency loss\\
                        $\Tilde \ell_e$ & computation loss \\
                        $\hat \ell_e$ & communication loss \\
                        $\bar \ell_e$ & storage loss \\
        
             $U$  & utility function \\
            $P$, &privacy budget function\\
                        $E$, &system efficiency function\\
            $F$ & aggregation method \\
            $w_{un}$ & unlearned model \\
            $w_{re}$ & retrained model \\
            $w_s$ & model of server  \\
            $w_c = \{w_1, \cdots, w_K\}$ &personalized model of $K$ clients  \\
            $w_g=\{w_s,w_c\}$ & unified model  \\
            
        $\mathcal C$ & contribution evaluation method \\
            $dist(\cdot, \cdot)$ & distance between two models \\       
            \hline
        \end{tabular}
    \label{tab:notation}
    \end{center}
\end{table}

\section{Problem 1: How to establish a foundational theory for FedFMs?}

\label{ftheory}

\subsection{Defining A Foundational Theory of FedFMs}

Although FedFMs hold great potential, they currently lack a solid theoretical foundation. 
This section defines the foundational theory of FedFMs as a multi-dimensional trade-off framework, focusing on balancing various goals such as privacy , utility,  efficiency, interpretability, fairness, robustness and aggregation methods. The aim is to emphasize the complex interactions between these dimensions and seek to achieve an optimal equilibrium that accommodates the diverse requirements of FedFMs.

The foundational theory of FedFMs involves three key trade-offs. First, the trade-off between privacy, utility, and efficiency. It provides a multi-objective optimization framework to ensure that improvements in one dimension do not compromise others, enabling the simultaneous optimization of all objectives in practical applications. Second, the trade-off between global and personalized model performance, which explores how to optimize the global model while maintaining flexibility for client-specific customizations. Finally, the trade-off between different types of FedFMs, which examines the differences between open-source (white-box) and closed-source (black-box) models, aiming to optimize the aggregation process for different model types and deployment scenarios.

\textbf{Problem Formulation.}
Consider \(K\) clients, each possessing private data \(D_k\), who participate in the FedFMs training process. Let \(w_s\) represent the model of the server, and \(w_c=\{w_1,\cdots,w_K\}\) denote the set of models of \(K\) clients. We define \(w_g = \{w_s, w_c\}\)\footnote{In this paper, the definition of $w_g,w_s$ and $w_c$ will be used consistently.
}.  We use $w$ as either $w_s, w_c$ or $w_g$. The foundation theory in FedFMs can be regarded as a multi-objective optimization \cite{kang2023optimizing,zhang2022no}, which aims to achieve a balance among privacy loss, utility loss, and efficiency loss as:
\begin{equation}
    \begin{split}
      &\min_{w}\Big( \ell_u(w), \ell_p(w), \ell_e(w) \Big), \\
\text{s.t. } &\ell_u(w) + \ell_p(w) + \ell_e(w) >0   
    \end{split}\label{eq:multi-objective}
\end{equation} 
$\text{where } \ell_i(w) = F\Big(\ell_{i,1}(w,D_1),\cdots, \ell_{i,K}(w,D_K))$ $ \text{for } i\in\{u,p,e\}$, $F$ is the aggregation mechanism such as FedAvg \cite{mcmahan2017communication}.  \(\ell_{u,k}(w,D_k)\), \(\ell_{p,k}(w,D_k)\), and \(\ell_{e,k}(w,D_k)\) denote the loss functions corresponding to the privacy, utility, and efficiency of client \(k\) for the data \(D_k\) respectively. Note that the sum of loss in Eq. \eqref{eq:multi-objective} for privacy, utility and efficiency is larger than zero, denoting the "no free lunch" aspect of the tradeoff \cite{zhang2022no}.

\subsection{Existing Methods}

\subsubsection{Multi-objective Trade-off Methods}

\begin{figure}[htbp]
    \centering
    \includegraphics[width=0.6\linewidth]{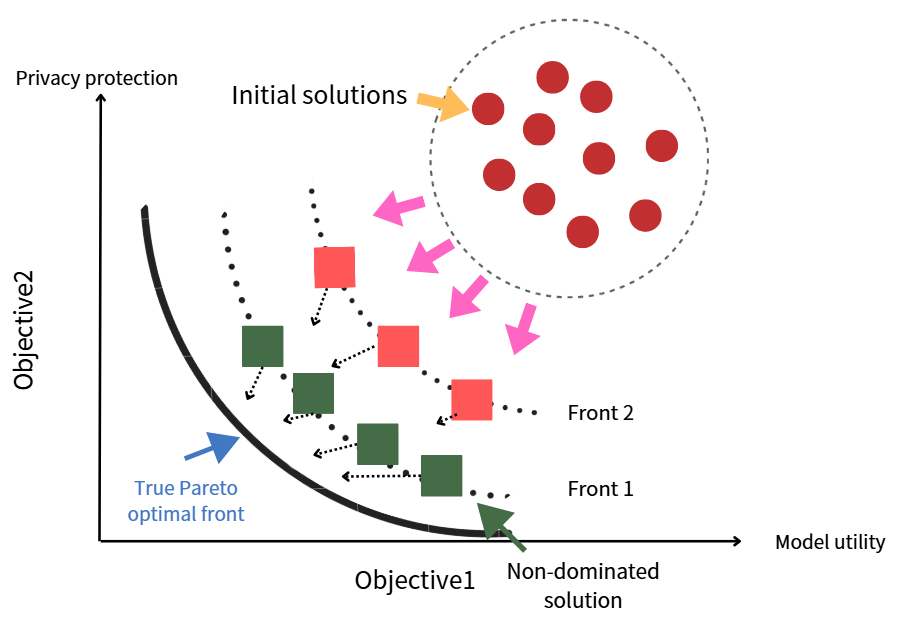}
    \captionsetup{justification=centering}
    \caption{Pareto Curve of Privacy-Utility Trade-off}
    \label{fig:Pareto}
\end{figure}

In previous FL research, multi-objective optimization problems are often represented as tasks where the “optimal balance point" lies on the Pareto front or surface, as illustrated in Fig. \ref{fig:Pareto}. This surface effectively identifies solutions that achieve the best balance, ensuring that improving one objective (e.g., privacy) does not come at the expense of others (e.g., utility or efficiency). 

First, we examine the trade-offs between privacy and other dimensions. In  FL, numerous studies have addressed the privacy-utility trade off~\cite{zhang2022no, kang2023optimizing}. For instance, \citet{zhang2022no} proposed the NFL theorem within the FL framework, indicating an inherent trade-off between privacy and utility, expressed as:
\begin{align}
C_1 \leq \ell_p + C_2\cdot\ell_u,
\end{align}
where $\ell_p $ represents the amount of privacy leakage, and $\ell_u$ represents utility loss. \citet{zhang2024no} further applied the NFL framework to LLMs. Unlike the privacy leakage defined by gradients exchanged between clients and servers in \cite{zhang2022no}, this framework focuses more on how much an attacker can infer the original data through protected embeddings. The core expression is:
\begin{align}
\label{eq:tv}
\frac{C_2}{C_1} \cdot \ell_p + \ell_u \geq C_2 \cdot \text{TV}(P \parallel \hat{P}),
\end{align}
where $\text{TV}(P \parallel \hat{P})$ represents the total variation distance between the distribution of the undistorted embedding and the distribution of the embedding independent of the client's prompt embedding. $\text{TV}(P \parallel \hat{P})$ is a constant that has nothing to do with the protection mechanism. Eq. (\ref{eq:tv}) indicates that when interacting with an LLM using protected prompts, it is impossible to minimize both privacy leakage and utility loss simultaneously.

Several studies have explored privacy-utility trade-offs in FedFMs. In FedFMs, privacy protection can be categorized into two scenarios: 1) absolute privacy protection, where no leakage is allowed, and 2) privacy protection with cost, where only specific data (e.g., names) is protected, focusing on key information. Common methods for absolute privacy protection include homomorphic encryption~\cite{zheng2024safely} and differential privacy~\cite{ hou2023privately}. For privacy with cost, techniques like knowledge distillation~\cite{wang2023can} and tunable soft prompts~\cite{dong2023tunable} can be applied.

Given the large number of parameters in FMs, optimizing efficiency without sacrificing performance is a critical challenge in FedFMs foundational theory. To this end, several methods~\cite{wu2024fedbiot,su2024titanic,yue2023fedjudge} have been proposed to enhance the efficiency of FedFMs systems. For instance, \citet{kuang2024federatedscope} introduced a federated parameter-efficient fine-tuning (PEFT) framework. Additionally, \citet{yue2023fedjudge} and \citet{su2024titanic} focused on selecting key parameters or subsets of clients to minimize communication costs, and \citet{wu2024fedbiot} leverages model compression for FedFMs.

In FedFMs, addressing system heterogeneity is crucial, as variations in data, resources, and personalization needs affect both global model generalization and client fairness. To tackle this, \citet{6-data-feddpa} employed global and local adapters to handle distribution shifts and personalized needs.

\subsubsection{Open- and Closed-Source FMs Trade-off Methods}

FMs can be categorized as closed-source (black-box) and open-source (white-box) based on parameter accessibility and usage rights. Closed-source FMs (e.g., GPT-4~\cite{OpenAI_GPT4_2023}) are proprietary, with their architecture, weights, and training processes kept confidential. These models are typically accessible only via APIs, limiting users' control over internal workings. In contrast, open-source FMs (e.g., LLaMA~\cite{touvron2023llama}) provide full access to architecture, weights, and training methods, enabling free access, modification, and redistribution by the community.

Aggregation plays a crucial role in training and fine-tuning FedFMs, requiring efficient knowledge integration while maintaining data privacy. However, aggregation methods differ between closed-source and open-source FMs, impacting how organizations deploy, train, and share models. These differences involve trade-offs in complexity, resource and feasibility.

In FedFMs, aggregation methods vary between open-source and closed-source models. For white-box FedFMs, where model weights and architectures are accessible~\cite{zhang2023towards, wang2024flora}, traditional parameter-based aggregation methods, such as FedAvg~\cite{mcmahan2017communication}, can be used. These methods allow for customized aggregation at different layers based on specific needs~\cite{zhang2023towards, li2020federated}. However, implementing open-source FedFMs requires significant AI expertise, computational resources, and infrastructure, which may not be feasible for all organizations. In contrast, black-box FedFMs, where model parameters are inaccessible~\cite{guo2023pfedprompt, yang2023efficient, bai2024diprompt}, focus on prompt-level aggregation. This involves aggregating high-level representations or prompt patterns~\cite{guo2023pfedprompt, yang2023efficient, bai2024diprompt}, rather than direct model parameters. This "prompt-centric" approach reduces technical barriers and operational overhead, making it accessible to organizations with limited AI expertise or resources, though it offers less customization than weight-based methods~\cite{yi2023pfedes}.

\subsection{Challenges and Potential Solutions}

\textit{Multi-objective optimization} in FedFMs remains challenging, as balancing privacy, utility, and efficiency across data heterogeneity, varying client resources, and network conditions is difficult. While Pareto optimization provides a theoretical framework, real-world complexities hinder achieving an optimal balance. Addressing \textit{fairness} across heterogeneous clients is also critical, as FedFMs rely on large-scale sensitive data, often highly heterogeneous, impacting system utility.
Furthermore, developing \textit{effective aggregation} methods for both open-source and closed-source models in FedFMs poses another challenge. Open-source models allow traditional parameter-based aggregation but require advanced AI expertise. Closed-source models accessed via APIs limit direct access to parameters and rely on less customizable prompt-level aggregation.

To address these challenges, potential solutions include: (1) Establishing \textit{theoretical bounds} for dimensions such as privacy, utility, and efficiency. (2) Adopting \textit{cross-layer optimization} strategies to balance the data, model, and communication layers. (3) \textit{Adaptive strategies} that dynamically adjust to client capabilities and network conditions, incorporating game theory or reinforcement learning, can optimize resource allocation and minimize communication overhead. (4) Developing \textit{hybrid aggregation} methods is crucial for overcoming the limitations of both open-source and closed-source aggregation approaches.
\section{Problem 2: How to Utilize Private Data in FedFMs?}

\subsection{Defining the Utilization of Private data in FedFMs}
\begin{figure}[]
    \centering
\includegraphics[width=1\linewidth]{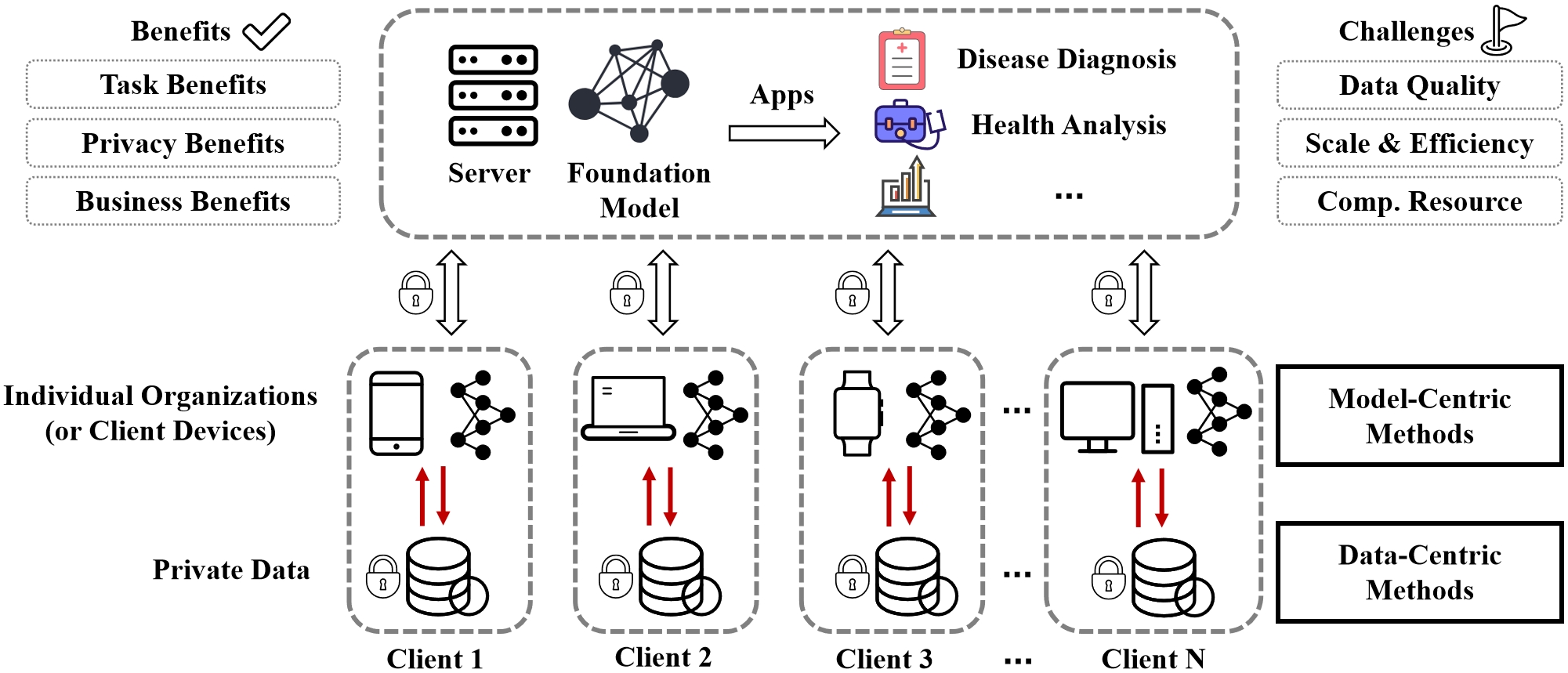}
    \caption{FedFMs utilize locally stored private data from organizations (toB) or client devices (toC) to train models while ensuring privacy, enabling applications like disease diagnosis and health analysis \cite{6-data-fedcampus}. }
    \label{fig:UtilPrivateData}
    \vspace{-5mm}
\end{figure}

Private data utilization in FedFMs enhances FMs by leveraging local private data while preserving privacy, as shown in Fig.~\ref{fig:UtilPrivateData}. In real-world scenarios, valuable private data is often isolated within organizations due to privacy regulations. For example, medical institutions maintain separate patient data repositories, creating “data islands” that limit training diversity. Traditional centralized training is impractical or restricted due to privacy concerns, which particularly affects FMs that rely on extensive, diverse datasets for robust performance. FedFMs overcome these challenges by allowing organizations to train models locally, sharing only model updates instead of raw data with a central server for aggregation. This approach effectively bridges data islands while ensuring strong privacy protections, addressing both data scarcity and regulatory constraints in FM development.

\textbf{Problem Formulation.} The utilization of private data in FedFMs involves clients contributing their private data for the model's training while preserving data privacy \cite{mcmahan2017communication,yang2019federated, 6-data-datasetgrouper}.
Consider a scenario where a central aggregation server and $K$ clients collaborate to learn a global FM. Each client is in a data-isolated environment, meaning that data is not shared among them. Let the $k$-th client possess data $D_k$. 
The optimization problem for obtaining the model $w_g$ with respect to data utilization can be formulated as:
\begin{equation}
    \begin{split}
       &\qquad \min_{w_g}\ell_u(w_g) \\
       & s.t.\quad \ell_{p,k}(w_g)<\delta_k, \text{for }k=1,\cdots,K
    \end{split}
\end{equation}
where $\ell_u(w_g) = F\Big(\ell_{u,1}(w_1), \cdots, {\ell}_{u,K}(w_K)\Big)$, $\ell_{u,k}(\cdot)$ represents the loss function related to data utilization for client $k$ and $F$ is the aggregation mechanism, such as FedAvg \cite{mcmahan2017communication}.  Additionally, $\ell_{p,k}(w)$ denotes the privacy leakage of client $k$ with respect to the model $w$, and it must be within a predefined threshold $\delta_k$.

\looseness=-1

\subsection{Existing Methods}
Existing methods for utilizing private data in FedFMs fall into two methods: \textit{data-centric} and \textit{model-centric}. Data-centric methods \textit{enhance distributed private data utility} while ensuring privacy, addressing data quality, efficiency, and performance. In contrast, model-centric methods adapt FM architectures and training paradigms to \textit{effectively leverage private data}, focusing on instruction tuning and model splitting.

\subsubsection{Data-Centric Methods }

\textbf{Data Quality Enhancement}. Distributed local privacy-protected data is often non-IID, varying in type, labels, and quality across clients, posing challenges for foundation models that require diverse, high-quality data. To address data imbalance, differential privacy-based synthetic data generation enhances utilization by creating balanced, privacy-preserving synthetic samples~\cite{6-data-dpsda}. Quality standardization mechanisms, such as training data scoring and global thresholds, ensure consistent data quality across clients~\cite{zhao2024enhancing}.\looseness=-1

\textbf{Computational Efficiency Enhancement}. 
Compared to centralized training, the non-IID nature of local privacy-protected data often increases communication rounds in federated learning, reducing training efficiency. To address this, FedHDS~\cite{qin2023federated} employs intelligent data selection to identify representative samples, reducing redundancy while preserving model quality. FedBPT~\cite{pmlr-v235-sun24j} leverages black-box APIs and gradient-free optimization to minimize data requirements without compromising performance.

\textbf{Privacy Risks Protection}.  
FMs remain vulnerable to privacy attacks, such as training data leakage, model parameter leakage, and architecture disclosure, which limit data diversity and scale. To address these risks, defensive frameworks like RoseAgg~\cite{6-data-roseagg} enhance protection against targeted attacks, including collusion and poisoning in federated settings. Practical implementations include FedFMS~\cite{6-data-fedfms} and FedKIM~\cite{6-data-fedkim}, which adapt foundation models for privacy-preserving medical applications.  FedCampus~\cite{6-data-fedcampus} and Fedkit~\cite{Fedkit}  provide comprehensive privacy solutions for smart campus environments.

\textbf{Balancing Global and Personalized Performance}. 
Effective private data utilization in FedFMs requires balancing global knowledge with personalized client insights, particularly when private data exhibits distinct patterns. For instance, in healthcare, integrating patient-specific data like genetics and lifestyle factors enhances diagnostic accuracy and personalized treatment \cite{6-data-health}, while global data provides broader coverage and general accuracy. Recent research has developed methods to optimize this trade-off: FedDPA~\cite{6-data-feddpa} uses specialized adapters for simultaneous global and local learning, and ZooPFL~\cite{lu2023zoopfl} employs zeroth-order optimization and client-specific embeddings for adaptation.

\subsubsection{Model-Centric Methods}

\textbf{Instruction Tuning}.
Traditional instruction tuning methods face challenges in federated settings due to the difficulty of obtaining high-quality, privately distributed instruction data. Recent solutions include: FedIT~\cite{zhang2024fedpit}, which leverages local device instructions and transfers multimodal knowledge at feature and decision levels to enhance adaptability; and OpenFedLLM~\cite{ye2024openfedllm}, which integrates instruction tuning and value alignment, demonstrating federated learning's superiority over local training across domains while preserving privacy.

\textbf{Model Splitting}.
The effective use of private data in FedFMs is hindered by computational constraints when processing local data. Split learning offers a solution: FedBone~\cite{JCST-2308-13639} splits the model into a cloud-based large-scale model and edge-based task-specific models, using conflicting gradient projection to improve cross-task generalization.
Extending to multimodal scenarios, M$^2$FedSA~\cite{zhangenhancing} modularizes models via split learning, retaining privacy-sensitive modules on clients and employing lightweight adapters to enhance task- and modality-specific knowledge.

\subsection{Challenges and Potential Solutions}

FMs face critical challenges due to their sensitivity to \textit{data quality issues}, which are more pronounced than in traditional FL, as existing quality control mechanisms are inadequate for their complex needs. Additionally, there is a \textit{scale and efficiency trade-off}, as FMs require vast amounts of high-quality data for training and tuning. Furthermore, \textit{computational resource constraints} are significant, as the cost of training FMs on private data far exceeds that of traditional models.

To address these challenges, potential solutions include:  
(1) \textit{Balancing data efficiency and computational cost} by designing unified data selection frameworks that use hierarchical sampling to identify valuable training instances, optimizing the trade-off between data usage and computational expenses.  
(2) \textit{Enhancing private data processing} through advanced federated split learning with adaptive layer-splitting strategies, tailoring model partitioning based on device capabilities, network conditions, and task requirements.  
(3) \textit{Refining data contribution evaluation} by establishing unified quality assessment metrics that consider data utility, task relevance, diversity, and their impact on existing model knowledge, ensuring more effective private data evaluation.

\section{Problem 3: How to Design Continual Learning in FedFMs?}
\label{cl}
\subsection{Defining Continual Learning in FedFMs}

\begin{figure}[htbp]
    \centering
    \includegraphics[width=0.9\linewidth]{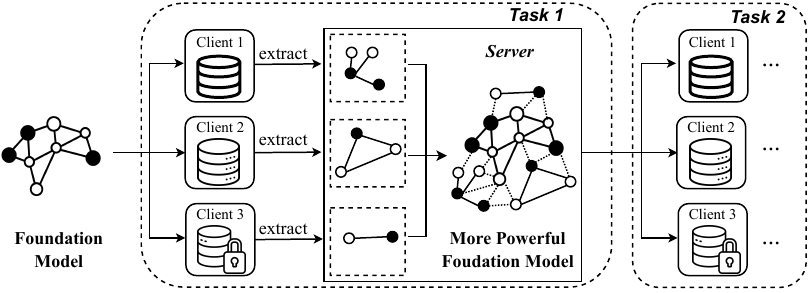}
    \caption{An illustration of Continual Learning with FedFMs. 
    }
    \label{fig:CLwithFM}
\end{figure}

Existing FL setups assume static models and data, neglecting the dynamic nature of real-world data collection. FedFMs, with their numerous parameters and complex training, struggle to meet real-time requirements without retraining. As the number of clients and data samples grows, FedFMs face scalability issues, potentially increasing communication overhead and slowing convergence. To address these, Continual Learning (CL) can be integrated into FedFMs for incremental updates without full retraining. Fig. \ref{fig:CLwithFM} illustrates our generalized FCL framework for FedFMs.

A crucial overlooked issue in FCL is the interaction between spatial and temporal heterogeneity, leading to spatial-temporal catastrophic forgetting (ST-CF) \cite{yang2024federatedcl}. Catastrophic Forgetting (CF) describes a deep model forgetting previous tasks after learning new ones, reducing accuracy \cite{kirkpatrick2017overcoming,li2017learning}. In FCL, clients experience temporal CF as they continually learn new tasks, while non-IID data causes spatial CF in the global model, decreasing performance on local test sets. Spatial forgetting interacts with temporal forgetting as clients use the global model for subsequent tasks \cite{yang2024federatedcl}.

The goals of FCL are threefold: \textbf{\textit{(1)}} Continuously update and optimize local models to adapt to changing local data. \textbf{\textit{(2)}} Enhance client knowledge through interaction with other clients via FL. \textbf{\textit{(3)}} Apply knowledge gained from other clients for further learning.

\textbf{Problem Formulation}.
This section extends the traditional FL framework to FCL to address the challenges posed by strong spatial-temporal data heterogeneity. 
Regarding spatial heterogeneity \cite{yoon2021federated,dong2022federated,ma2022continual}. $K$ clients have distinct data denoted as $D_k$. For temporal heterogeneity, in the federated system, clients are involved in learning $m$ tasks. The corresponding datasets for each client $k$ are $\{D_k^1,\ldots,D_k^{m}\}$.

When all clients have collaboratively completed learning task $1, \cdots, T$ and are about to embark on learning the new task $T + 1$, the goal of federated continual learning encompasses two key aspects. Firstly, it is to effectively learn the new task $T+1$. Secondly, it is to ensure that knowledge of the previous tasks $1,\cdots,T$ is not forgotten. Specifically, the optimization objective is formulated as:
\begin{equation}
\begin{split}
    \min_{w_g}F\Big(\underbrace{\ell_{u,1}(w_g,\{D_1^t\}_{t = 1}^T), \cdots, \ell_{u,K}(w_g,\{D_K^t\}_{t = 1}^T)}_{\text{old task loss}}, \\ \underbrace{\ell_{u,1}(w_g,D_1^{T + 1}),\ell_{u,K}(w_g,D_K^{T + 1})}_{\text{new task loss}}\Big),
    \end{split}
\end{equation}
where $\ell_{u,k}(w_g,\{D_k^t\}_{t = 1}^T)$ and $\ell_{u,k}(w_g,D_k^{T + 1})$ are the loss functions of client $k$ corresponding to the old tasks and the new task, respectively. $F$ is the aggregation mechanism.

\subsection{Existing Methods}

The keys to the FCL problems are how to integrate the heterogeneous knowledge from different clients, and how the clients can utilize this fused knowledge for future learning. Inspired by the tri-level (input, model-parameter and output) division in class-incremental learning \cite{zhou2024class} and existing survey of FCL \cite{yang2024federatedcl,wang2024federated}, we categorize the most effective and well-known existing methods into the following types: (1). \textit{Replay-based Methods}, (2). \textit{Methods based on Regularization and Decomposition
}, (3). \textit{Distillation-based Methods} and (4). \textit{Prompted-based Methods}.

\textbf{Replay-based Methods} involve either retaining samples in their original form or creating synthetic samples using a generative model. These samples from previous tasks are reintroduced during the learning of new tasks to mitigate forgetting. \citet{li2024towards,li2024sr} cached selected previous samples based on their global and local importance. 
As for the generative ways, \citet{yu2024overcoming} (FedCBC) built class-specific variational autoencoders for each class on the client side, performing classification tasks through anomaly detection, thus avoiding forgetting caused by interference and overlap between classes. \citet{liang2025diffusion} utilized a pre-trained conditional diffusion model to deduce class-specific input conditions within the model's input space, substantially cutting down on computational resources while ensuring effective generation.

\textbf{Methods based on Regularization and Decomposition} are to add appropriate regularization terms to model parameters during optimization to preserve old knowledge, or to divide the model into multiple parts to prevent forgetting.
\citet{yoon2021federated} (FedWeIT) decomposed parameters into local, global-based and task-adaptive parts to address both practical and pathological data heterogeneity in FCL. \citet{dong2022federated} (GLFC) utilized a class-aware gradient compensation loss and a class-semantic relation distillation loss to mitigate forgetting, and a proxy server to alleviate data heterogeneity.

\textbf{Distillation-based Methods} are a popular technique to fuse knowledge, the core idea is to distill the knowledge contained in an already trained teacher model into a student model \cite{hinton2015distilling}. \citet{ma2022continual} introduced CFeD, a method that employs knowledge distillation at both client and server levels in continual federated learning, with each client having a separate unlabeled dataset to reduce forgetting and leverage unused computational resources. 
\citet{wei2022knowledge} (FedKL) separated the training goal into two parts: classification and knowledge retention. In the knowledge retention part, classes not available locally are supervised through global model distillation with logistic regression loss.

\textbf{\textit{Prompted-based Methods}} leverage the pre-trained Vision Transformer (ViT) \cite{dosovitskiy2020image}, which has strong representation capabilities, by adding corresponding prompts or fine-tuning it to enhance its performance on downstream tasks. To our knowledge, this is the only paradigm closest to the utilization of FedFMs so far. \citet{piaofederated} proposed POWDER 
to effectively foster the transfer of knowledge encapsulated in prompts between various sequentially learned tasks and clients. \citet{yu2024personalized} proposed a novel concept called multi-granularity prompt (FedMGP), i.e., coarse-grained global prompt acquired through the ViT learning process, and fine-grained local prompt used to personalize the generalized representation.

\subsection{Challenges and Potential Solutions}

\textit{Spatial-temporal data heterogeneity} is a significant challenge, with data varying across clients and within different tasks of the same client, leading to spatial-temporal catastrophic forgetting \cite{yang2024federatedcl}.
FedFMs must address \textit{knowledge transfer and error correction} during federated continual learning (FCL), leveraging knowledge from other clients to learn future knowledge and using later-acquired knowledge to correct past errors. Furthermore, \textit{knowledge conflict} arises due to the complexity in real-world applications, where knowledge between clients or at different times can be incompatible or conflicting.

To address these challenges,  potential solutions include: (1) Utilizing an \textit{external knowledge base} to provide pre-existing knowledge, enriching the FedFMs with additional context and information. (2) Implementing \textit{selectively knowledge fusion} within FedFMs to prevent conflicts and identify malicious clients, ensuring robust and secure model updates.

\section{Problem 4: How to Design Unlearning in FedFMs?}

\subsection{Defining Unlearning in FedFMs}

Data is crucial for intelligent development across industries, but data leakage risks during circulation and sharing have raised privacy concerns. Global governments and legislators have imposed stringent data privacy regulations, such as GDPR and CCPA, mandating digital service providers to grant users the ``Right to Be Forgotten" (RTBF) \cite{dang2021right} and establish mechanisms for data removal from models \cite{liu2022right}. 

This opt-out mechanism, known as machine unlearning (MU) \cite{Gu2024Unlearning}, is vital in FedFMs. Despite FL not requiring direct user data access, it retains implicit information from training, posing privacy risks from model inversion attacks \cite{ren2022grnn}. Additionally, training data may include unauthorized or biased content, raising ethical and copyright issues \cite{Carlini2023Quantifying}. Addressing these concerns is essential for responsible FedFM deployment. Due to the high retraining cost, developing a federated unlearning (FU) technique that aligns with legal frameworks is a key challenge for FedFM advancement. 

\begin{figure}[ht]
    \centering
     \includegraphics[width=0.49\textwidth]{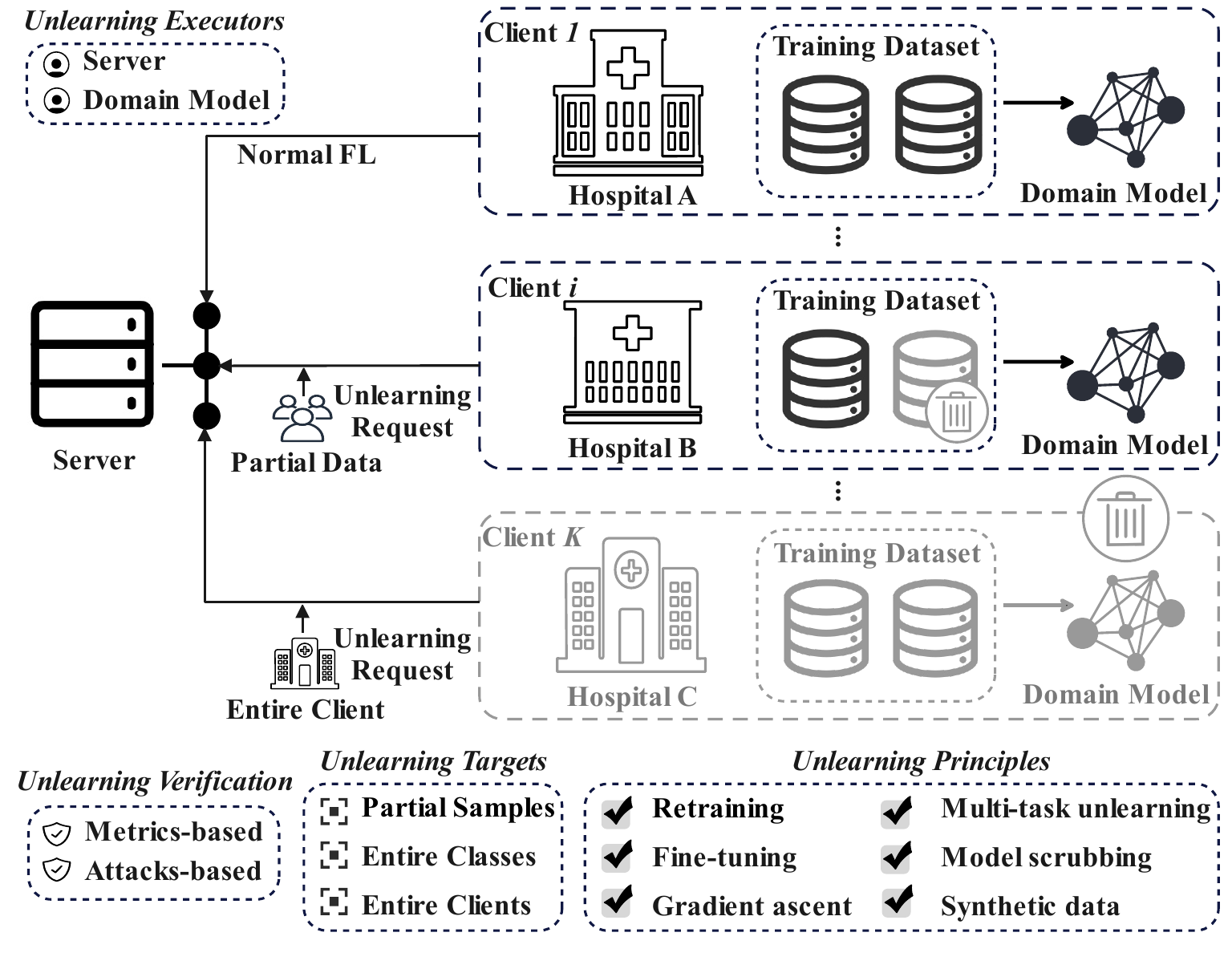}
     \caption{Machine Unlearning in FedFMs.}
     \label{fig: 5-Unlearning-framework.pdf}
     \vspace{-5mm}
 \end{figure}

The unlearning process in FedFMs, illustrated in Fig. \ref{fig: 5-Unlearning-framework.pdf}, involves identifying and flagging clients and data records associated with data deletion requests, then applying active forgetting techniques to remove specified data contributions while preserving model integrity.

\textbf{Problem Formulation.}
The objective of FU is to mitigate the impact of a specific subset of data, denoted as $\tilde{D}_{j}$, belonging to the target client $j$ \cite{wang2022federated, shah2023unlearning,Gu2024Unlearning}. 

In the context of FU, we need to construct a new model $w_{un}$. To do this, we first define a reference model $w_{re}$. The model $w_{re}$ is obtained by minimizing a weighted sum of loss functions. Here, $F$ is the aggregation mechanishm such as FedAvg \cite{mcmahan2017communication} . The model $w_{re}$ is given by:
\begin{equation}
\begin{split}
 w_{re}=\argmin_w F \Big (\ell_1(w, D_1). \cdots, \ell_{j-1}(w, D_{j-1}), \\
\underbrace{\ell_j(w, D_j - \tilde{D}_{j})}_{\text{removing data}\tilde{D}_{j}}, \cdots, \ell_K(w, D_K - \tilde{D}_{K})\Big)   
\end{split}
\end{equation}
Formally, the model $w_{un}$ constructed by the unlearning algorithm should satisfy the following optimization objective:
\begin{equation}\label{eq:unlearn}
\min_{w_{un}}\text{dist}\Big(\underbrace{w_{un}}_{\text{unlearned model}}, \underbrace{w_{re}}_{\text{retrained model}}\Big)
\end{equation}
where $\text{dist}(\cdot,\cdot)$ represents a measure of the distance between two models. Common choices for this distance measure include the Euclidean distance or cosine similarity \cite{gu2024ferrari}, KL divergence \cite{van2014renyi} and Maximum Mean Discrepancy \cite{pan2008transfer}. By minimizing this distance, Eq. \eqref{eq:unlearn} ensures that the model $w_{un}$ closely approximates the retrained model $w_{re}$.

\subsection{Existing Methods}
We summarize the four key aspects of the FU process: unlearning targets, unlearning executors, unlearning verification, and unlearning principles.

\subsubsection{Unlearning Targets}
Unlearning requests are typically classified into four types:\textbf{ partial samples}, \textbf{entire classes}, \textbf{sensitive features}, and \textbf{entire clients}. Partial sample unlearning removes specific data samples' contributions, representing fine-grained unlearning \cite{ma2022learn, shah2023unlearning}. Feature unlearning eliminates specific data features, such as facial attributes \cite{gu2024ferrari}. Class unlearning removes data contributions from one or more classes, like forgetting facial images of a specific user in facial recognition models \cite{wang2022federated, gu2024few}. Client unlearning involves revoking all data contributions from a client, which can lead to catastrophic forgetting due to the large-scale data removal \cite{Gu2024Unlearning, nguyen2022markov}.

\subsubsection{Unlearning Executors}

Unlearning executors include server-side and domain model-based approaches.\textbf{ Server-side unlearning} adjusts global model parameters, initially requiring full retraining, which is costly in federated settings \cite{liu2022right}. Recent methods focus on eliminating target user influence while restoring model performance through relearning and parameter adjustments \cite{su2023asynchronous, fraboni2024sifu}. \textbf{Domain model-based unlearning} uses additional training to update the global model without full retraining, employing techniques like training update correction and gradient correction, which minimize complexity and resource overhead \cite{hacohen2019power, wu2022federated, gao2024verifi, halimi2022federated}. These approaches typically preserve higher model accuracy than server-side methods.

\subsubsection{Unlearning Verification}
Unlearning verification in FU aims to assess the effectiveness, efficiency, and security of data removal from models, typically categorized into metric-based and attack-based approaches. \textbf{Metric-based verification} evaluates model performance using accuracy, loss, and statistical errors on target and test datasets to ensure proper data removal and model robustness. Model discrepancy is measured using techniques such as Euclidean distance or KL-divergence, while execution efficiency is assessed through metrics like time, memory usage, and speed-up ratios \cite{cao2023fedrecover}. \textbf{Attack-based verification} simulates adversarial scenarios, such as membership inference and backdoor attacks, to test if unlearned data can still be inferred or if malicious triggers persist. Reduced attack success rates indicate effective unlearning \cite{zhang2023fedrecovery}.

\subsubsection{Unlearning Principles}

 Existing unlearning methods rely on various principles to align the distribution of the modified model $ W $ with that of a retrained model $ w_{re} $ \cite{liu2024survey}. \textbf{Retraining} removes all influence of $ D_u $ by training from scratch on $ D_r $, but it is resource-intensive. \textbf{Fine-tuning} optimizes $ W $ on $ D_r $ to reduce the impact of $ D_u $, though it requires multiple iterations. \textbf{Gradient ascent} maximizes loss to reverse the influence of $ D_u $, but risks catastrophic forgetting unless constraints are applied. \textbf{Multi-task unlearning} balances the removal of $ D_u $'s influence with reinforcing $ D_r $'s knowledge. \textbf{Model scrubbing} transforms $ W $ using a quadratic loss approximation to mimic retraining, but the computational cost of approximating the Hessian limits scalability. Finally, \textbf{synthetic data} replaces or combines $ D_u $ with synthetic data to disentangle its influence while preserving model performance. 
 Each method involves trade-offs between effectiveness, efficiency, and scalability.
 \citet{shao2024federated} further analyzed unlearning in federated settings, revealing additional side-effects like stability and fairness stem from data heterogeneity.

\subsection{Challenges and Potential Solutions}

FedFMs encounter additional unlearning challenges due to \textit{high model complexity}, \textit{knowledge coupling}, and \textit{cross-client consistency} issues. High model complexity complicates targeted updates, risking model integrity and performance. Knowledge coupling in multi-modal FedFMs necessitates comprehensive eradication of associated knowledge. Maintaining cross-client consistency during unlearning is challenging, requiring robust synchronization and validation.

To address these challenges, potential solutions include: (1) \textit{Modular Optimization Unlearning}, where modular architectures localize unlearning to specific components, using techniques like constrained fine-tuning and sensitivity analyses to identify influenced parameters. (2) \textit{Disentangled Knowledge Decoupling} separates knowledge across modalities, enabling targeted unlearning with techniques like cross-modal attention tracking and mutual information minimization. (3) \textit{Federated Cross-Client Coordination} ensures consistency through federated consensus mechanisms, using gradient averaging and replay buffers to synchronize updates and align with unlearning objectives.
\section{Problem 5: How to address NON-IID issues and graph data in FedFMs}

\subsection{Defining the Non-IID Issues and Graph Data in FedFMs}

The Non-IID issues stem from heterogeneous data distributions across clients, where each client may have distinct feature sets, label distributions, and sample sizes. These variations pose significant challenges to global model training, particularly in FedFMs~\cite{ren2024advances,zhong2023semi, Yang_Zhang_Dai_Pan_2020}. The primary types of Non-IID data are outlined as follows: 1) \textbf{Feature distribution Skew}, where clients have distinct feature distributions, leading to data-specific patterns; \textit{e.g}., hospitals serving different age groups. 2) \textbf{Label distribution Skew}, where differing label distributions hinder model generalization, as seen in transaction classification with varying data types; 3) \textbf{Quantity Skew}, where data volume disparities across clients bias aggregation, reducing model representativeness. 4) \textbf{Temporal Skew}, when clients collect data at different times, complicating model generalization, such as regional weather sensors recording at non-synchronized times.

Beyond these static skews, dynamic shifts like concept drift (same label, different features) and concept shift (same features, different label) introduce additional complexity. Concept drift occurs when $P_k(x | y)$ changes while $P(y)$ remains consistent, as seen in regional variations of "winter" images. Concept shift arises when $P_k(y | x)$ varies despite a stable $P(x)$, such as differing labels for similar behaviors across cultures.

\begin{figure}[htbp]
    \centering
    \includegraphics[width=0.95\linewidth]{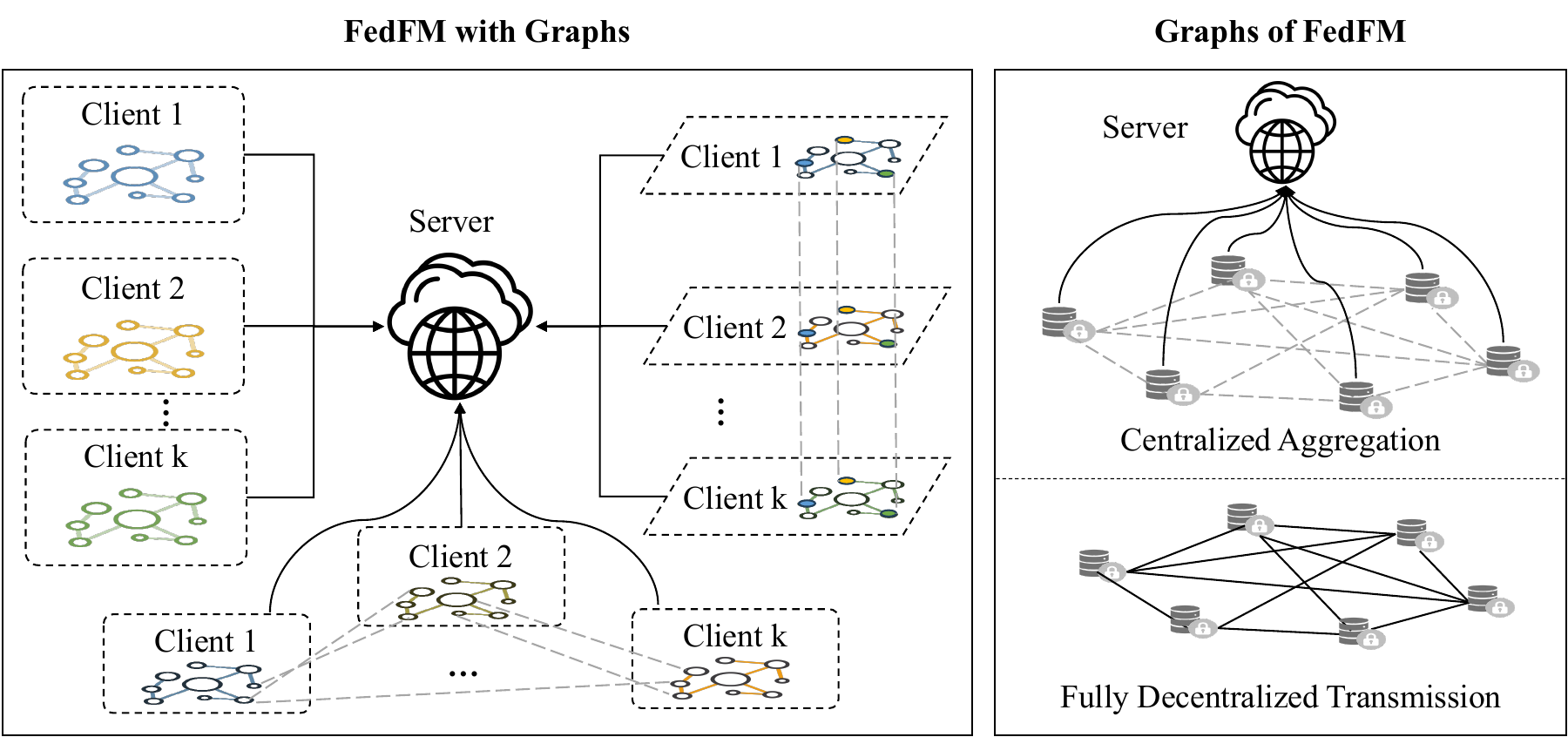}
    \caption{FedFMs with Graphs and Graphs of FedFMs.}
    \label{fig:Graph-FM}
    \vspace{-3mm}
\end{figure}

The federated graph of models in FedFMs presents two key perspectives: \textbf{FedFMs with Graphs} and \textbf{Graphs of FedFMs}, as shown in Figure~\ref{fig:Graph-FM}. In the former, each client’s data forms graph structures capturing internal and external dependencies, requiring models to learn both intra- and inter-graph relationships~\cite{zhang2021federated}. In the latter, clients act as nodes in a federated network, with edges representing relationships like geographic proximity or data similarity~\cite{fu2022federated}. This \textit{Internet of Models} enables selective information sharing under Non-IID conditions but demands advanced aggregation strategies to ensure model consistency.

\textbf{Problem Formulation.} There are two different settings in this section involving a server and multiple clients: \textit{Non-IID issues} and \textit{FedFMs with Graphs}.

\begin{itemize}
  \item  \textbf{Setting 1. Non-IID issues}. In the FedFMs framework \cite{zhang2021federated,fu2022federated}, the optimization problem with Non-IID data across $K$ clients is formulated as follows:
\begin{equation}
\begin{aligned}  
&\min_{w_g}F\Big(\ell_{u,1}(w_g,D_1), \cdots, \ell_{u,K}(w_g,D_K) \Big) \\  
&\text{where}\quad dist(D_{k_1}, D_{k_2})>0\ \forall k_1\neq k_2
\end{aligned}
\end{equation}
where \(\ell_{u,k}(\cdot)\) represents the utility loss function of client \(k\), $F$ is the aggregation mechanishm such as FedAvg and $\text{dist}(\cdot, \cdot)$ represent the distance of two dataset, \textit{e.g.},  cosine similarity, KL divergence and Maximum Mean Discrepancy.

  \item \textbf{Setting 2. FedFMs with Graphs}. In this setting, each client \(k\) possesses a local graph \(G_k=(V_k, E_k)\). The set \(V_k\) contains the nodes, and \(E_k\) represents the set of edges within the local graph. $K$ clients consist the whole graph as $G =(V ,E)$, where vertices are $K$ nodes \cite{baek2023personalized}. The objective of learning a federated graph neural network (GNN) model \(w_g\) is given by:
\begin{equation}\label{eq:graph}
\begin{split}
\min_{w_g}&F\Big(\ell_{u,1}(w_g, G_1, G), \cdots, \ell_{u,K}(w_g, G_K, G) \Big)\\
& s.t.\quad \ell_{p,k}(w_g, G_k, G)<\delta_k, \text{for }k=1,\cdots,K
    \end{split}
\end{equation}
where \(\ell_{u,k}(\cdot)\) is the utility loss function of client \(k\). Additionally, $\ell_{p,k}(\cdot)$ denotes the privacy leakage of client $k$, $F$ is the aggregation mechanishm and it must be within a predefined threshold $\delta_k$. Compared to the Non-IID problem, Eq. \eqref{eq:graph} considers the graph data $G$ and $G_k$.
\end{itemize}

\subsection{Existing Methods}

To address Non-IID challenges in FedFMs, techniques like knowledge distillation and model pruning unify models into a homogeneous representation, reducing optimization complexity~ \cite{fan2023fate,lachi2024graphfmscalableframeworkmultigraph}. Alternatively, modeling model relationships as a graph optimization problem, where client models are nodes and edges represent feature similarity, task relevance, or proximity, enhances adaptability in federated systems~\cite{yu2024netsafe, feng2024graphroutergraphbasedrouterllm}. Existing methods address specific aspects of FL challenges as follows:

\textbf{Distribution Adaptation}. These methods enhance global model performance in Non-IID settings by addressing distributional shifts. \citet{park2024fedbaffederatedlearningaggregation} (FedBaF) biases aggregation with a pre-trained foundation model to leverage its generalization capabilities. \citet{imteaj2024tripleplayenhancingfederatedlearning} (TriplePlay) integrates foundation models, using CLIP as an adapter to mitigate data heterogeneity. \citet{6-data-feddpa} (FedDPA) employs a dual-adapter framework with dynamic weighting for test-time adaptation and personalization. These approaches collectively improve FedFMs' adaptability to diverse client distributions.

\textbf{Federated Graph of Models}. Such work is typically related to \textit{Federated Graph Foundation Models with Topology Optimization}, focusing on scalable learning across distributed graph-structured data by optimizing client relationships for efficient information sharing. \citet{li2024fedgtatopologyawareaveragingfederated} (FedGTA) enhances federated graph learning with topology-aware local smoothing and mixed neighbor features. \citet{ma2024beyond} proposed \textit{client topology learning} and \textit{learning on client topology}, leveraging client topology to train models robust to out-of-federation data. \citet{huang2022accelerating} introduced TOFEL, optimizing aggregation topology and computing speed to reduce energy consumption and latency. These approaches enhance performance, scalability, and communication efficiency in graph-based FedFMs.

\textbf{Optimization in Graph of Models}. Achieving a sustainable balance between security, fairness, and personalization in model networks is crucial for adapting to dynamic environments while ensuring stable performance. A key challenge in federated topologies is preserving model privacy. \citet{chen2024privfusion} (PrivFusion) introduced a privacy-preserving fusion architecture using graph structures and hybrid local differential privacy. \citet{chen2024model} proposed a cross-federation model fusion framework to address privacy, population shifts, and fairness in global healthcare. Beyond privacy, complexity and personalization are also critical.

\subsection{Challenges and Potential Solutions}

\textit{Convergence issues} from heterogeneous client data hinder optimization, causing slow or unstable convergence, especially in large-scale federated systems. The \textit{complexity of dynamic network topologies} complicates communication, aggregation, and information flow, exacerbated by computational challenges. In the Internet of Models paradigm, \textit{sustainable optimization} requires balancing utility, privacy, fairness, and personalization, further complicated by diverse node priorities.

To address these challenges, potential solutions include: (1) \textit{Adaptive Optimization} through personalized FL algorithms and hierarchical clustering to align local and global objectives. (2) \textit{Dynamic and Topology-aware Aggregation} to manage evolving network structures and ensure effective communication. (3) \textit{Multi-objective Optimization}, integrating differential privacy, fair allocation, and reinforcement learning to enhance scalability, adaptability, and fairness in Non-IID and graph-based environments.

\section{Problem 6: 
How to achieve the bidirectional knowledge transfer between FMs and DMs in FedFMs?
}

\subsection{Defining Bidirectional Knowledge Transfer in FedFMs}

Bidirectional knowledge transfer within FedFMs denotes the reciprocal exchange and adaptation of knowledge between the server-hosted FMs and the client-owned DMs. This dynamic interplay entails not only the transfer of knowledge from the FM to the DMs to bolster their capacities but also the converse flow of domain-specific insights from the DMs back to the FMs, thereby enriching its comprehension and performance across a spectrum of domains. 
This interaction between the server-based FMs and client-based small DMs is analogous to the teaching and learning interactions between a teacher and students; knowledge flows both ways while privacy is preserved.

\begin{figure}[ht]
    \centering
     \includegraphics[width=0.45\textwidth]{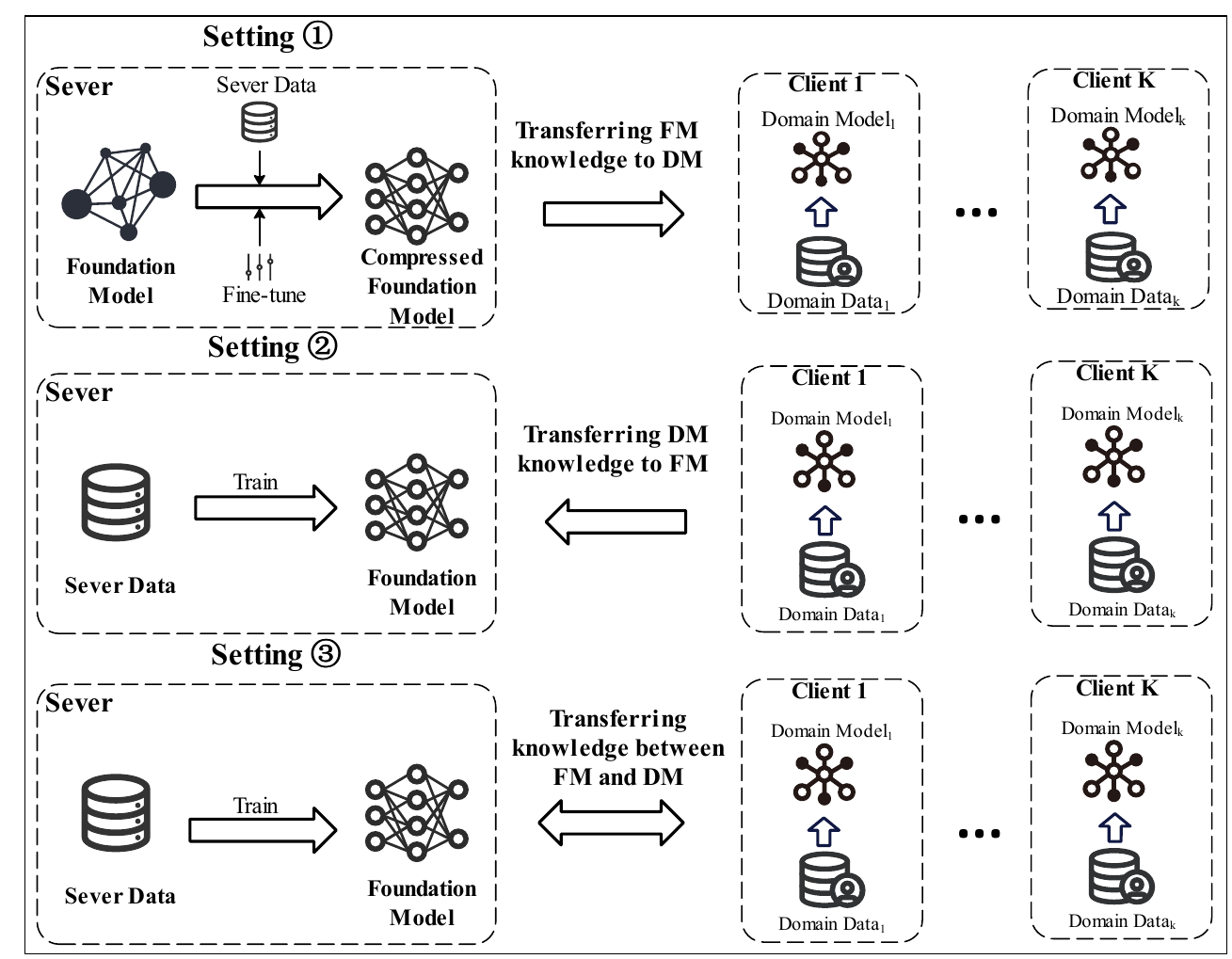}
      
     \caption{Illustration of the three settings for bidirectional knowledge transfer in FedFMs.}
     \label{fig:Bidirectional}
 \end{figure}

The framework of bidirectional knowledge transfer can be delineated into three distinct settings \cite{kang2023grounding,xiao2023offsite,deng2023mutual}, as illustrated in Fig. \ref{fig:Bidirectional}.  \textit{Setting 1} aims to facilitate the transfer and adaptation of knowledge from the server's FM to the clients' DMs. \textit{Setting 2} focuses on harnessing the domain-specific expertise of clients to refine the server's FM. \textit{Setting 3} strives for the co-optimization of both the FM and the DMs, fostering a synergistic enhancement of their respective capabilities. 

\textbf{Problem Formulation}. 
There are three different settings in this section involving a server and multiple clients. Here, \(w_s\) represents the server's FM, \(w_k\) represents the DM of client \(k\), \(D_k\) is the local private dataset of client \(k\), and \(D_p\) is the public data available at the server.

\begin{itemize}
\item \textbf{Setting 1: Optimizing clients' DMs using server's FM and local data.}
The goal is to optimize the DMs (\(w_k\)) of the client $k$ by taking advantage of the knowledge of the server's model \(w_s\) and the local private dataset (\(D_k\)) of each client. The main objective is formulated as:
\begin{equation*}\label{eq:bi - setting_1}
\min_{w_1, \cdots, w_K}F\Big(\ell_{u,1}(w_1|w_s ,D_1), \cdots, \ell_{u,K}(w_K|w_s ,D_K) \Big),
\end{equation*}
where \(\ell_{u,k}(w_k|w_s ,D_k)\) represents the utility loss function of client \(k\).

\item \textbf{Setting 2: Optimizing the server's FM using clients' domain-specific knowledge and public data.}
    This setting aims to optimize the server's FM (\(w_s\)) by leveraging the domain-specific knowledge of the clients' models (\(w_{k}\), where \(k\in\{1,\cdots,K\}\)) and the public data (\(D_p\)). The main objective is given by:
\begin{equation*}\label{eq:bi - setting_2}
\min_{w_s}F\Big(\ell_{u,1}(w_s|w_1 ,D_P), \cdots, \ell_{u,K}(w_s|w_K ,D_P) \Big),\end{equation*}
where \(\ell_{u,s}(w_s|w_k, D_p)\) is the utility loss function of the server. 

    \item \textbf{Setting 3: Co-optimizing the server's FM and clients' DMs.}
    This setting seeks to co-optimize the server's FM (\(w_s\)) and the clients' DMs (\(w_k\) for \(k\in\{1,\cdots,K\}\)) by mutually leveraging the knowledge of each other. The main objective is formulated as:
\begin{equation}\label{eq:bi - setting_3}
\begin{split}
    \min_{w_s, w_1, \cdots, w_K}\alpha F_1\Big(\ell_{u,1}(w_1|w_s ,D_1), \cdots, \\\ell_{u,K}(w_K|w_s ,D_K) \Big)\\ + (1-\alpha)F_2\Big(\ell_{u,1}(w_s|w_1 ,D_P), \cdots, \\\ell_{u,K}(w_s|w_K ,D_P) \Big),
    \end{split}
\end{equation}
where \(\ell_{u,k}(w_k|w_s ,D_k)\) is the utility loss function of client \(k\), and \(\ell_{u,s}(w_s|w_k, D_p)\) is the utility loss function of the server. $F_1$ and $F_2$ are the one of the aggregation methods as $F$. $\alpha$ is the coefficient ranging from 0 to 1.
\end{itemize}

\subsection{Existing Methods}
\subsubsection{Setting 1 - Transferring knowledge from server's FM to downstream clients' DMs}

Existing methods encompass data-level, representation-level, and model-level knowledge transfer.

In data-level knowledge transfer, works focus on knowledge distillation and synthetic data utilization. \citet{hsieh2023distilling} proposed DSS, a knowledge distillation framework based on Chain of Thought (COT) for distilling LLMs step-by-step. \citet{li2022explanations} introduced MT-COT, leveraging LLM-generated explanations to enhance smaller reasoning models via multitask learning. \citet{jiang2023lion} presented Lion, an adversarial distillation framework for efficient LLM knowledge transfer. \citet{fan2024pdss} developed PDSS, a privacy-preserving framework for distilling step-by-step FMs using perturbed prompts and rationales. \citet{li2024federated} proposed FDKT, a federated domain-specific knowledge transfer framework utilizing a generative pipeline on private data with differential privacy guarantees.

In representation-level knowledge transfer, works concentrate on split learning and knowledge distillation. \citet{shen2023split} investigated the impact of the split position in the FM on privacy-preserving capacity and model performance in the SAP framework. \citet{he2019model} introduced FedGKT, a method tailoring the server's FM to clients' DMs using knowledge distillation between the FM and DM. \citet{gu2024minillm} presented MINILLM, which distills LLMs in a white-box setting by minimizing reverse Kullback-Leibler divergence.

In model-level knowledge transfer, \citet{yu2023selective} proposed SPT, a selective pre-training approach that pre-trains a FM on training data chosen by a domain classifier trained on clients' private data via DP-SGD. \citet{zhang2023gpt} utilized FMs to generate synthetic data for training a model, which is then distributed to clients for initialization and fine-tuning with private data within the standard FL framework.

\subsubsection{Setting 2 - Enhancing  server's FM with domain knowledge from downstream clients}

Existing methods for enhancing the server's FM with domain knowledge from downstream clients focus on representation-level and model-level knowledge transfer. 

In representation-level knowledge transfer, \citet{yu2023multimodal} proposed CreamFL, a multimodal FL framework based on knowledge distillation, which aggregates representations to facilitate knowledge transfer. CreamFL uses a global-local cross-modal aggregation strategy and inter/intra-modal contrastive objectives. On the server side, the FM is trained via knowledge distillation using client-provided aggregated representations.

In model-level knowledge transfer, \citet{xiao2023offsite} introduced Offsite-Tuning, a privacy-preserving and efficient transfer learning framework. It splits the FM into a trainable adapter and a frozen emulator. Clients fine-tune the adapter on their data with the emulator's help and send it back to the server for integration. \citet{fan2023fate} extended this framework to FL, where the server distributes the emulator and adapter to clients, who fine-tune the adapter and return it. The server aggregates the adapters securely and integrates them into the FM.

\subsubsection{Setting 3 - Co-optimizing both server’s FM and clients’ DMs}

Approaches for co-optimizing the server’s FM and clients’ DMs also involve representation-level and model-level knowledge transfer. 

In representation-level knowledge transfer, \citet{fan-etal-2025-fedmkt} introduced FedMKT, a federated mutual knowledge transfer framework that uses selective knowledge distillation and token alignment (via MinED) to enhance both the FM and DMs, addressing model heterogeneity.

In model-level knowledge transfer, \citet{fan2024fedcollm} proposed FedCoLLM, a parameter-efficient federated framework using lightweight LoRA adapters alongside DMs for privacy-preserving knowledge exchange. It also employs mutual knowledge distillation between the FM and aggregated DM on the server. \citet{deng2023mutual} introduced CrossLM, a framework for federated learning and mutual enhancement of client-side DMs and a server-side FM without sharing private data. DMs guide the FM to generate synthetic data, which is validated by clients' DMs, improving both models and enabling data-free knowledge transfer.

\subsection{Challenges and Potential Solutions}

In the realm of bidirectional knowledge transfer, challenges such as \textit{data heterogeneity}, \textit{representation heterogeneity}, \textit{model heterogeneity}, and \textit{privacy concerns} persist. 

To address these challenges, potential solutions include: (1) \textit{Unified Model Architectures} to promote standardization and compatibility. (2) \textit{Adaptive Knowledge Transfer} methods that dynamically adjust based on model needs and capabilities \cite{jiang2023lion}. (3) \textit{Synthetic Data} to augment training processes without sole reliance on real-world data. (4) \textit{Advanced Privacy Techniques} like differential privacy \cite{dwork2006differential}, homomorphic encryption \cite{gentry2009fully}, and secure multi-party computation \cite{yao1986generate} to mitigate privacy risks without compromising performance.

\section{Problem 7: How to design incentive mechanisms through contribution evaluation in FedFMs?}
\label{con_eva}

\subsection{Defining Incentives through Contribution Evaluation}

    As the scaling law shows, the FMs' performance relies on high-quality data and substantial computational resources, which need appropriate compensation to be contributed in the real-world.
    Thus, it is crucial for FedFMs to attract data providers and incentive them to contribute high-quality data or other resources.
    The contribution evaluation for FedFMs aims to enable the client's participation by measuring the importance of datasets or computational resources.
    The incentive mechanisms for FedFMs aim to ensure the data providers' active and sustained participation by providing them with fair rewards.
    Contribution evaluation in FedFMs refers to measuring the importance or influence of heterogeneous datasets, model quality, or computational resources, which is vital for identifying the free riders and outliers within the FedFMs ecosystem and can serve as a prerequisite for the incentive mechanisms in FedFMs.

\textbf{Problem Formulation}.
Consider a scenario with \(K\)  clients, where each client \(k\) has a private dataset \(D_{k}\). A contribution evaluation mechanism $\mathcal C$ is designed to compute the contribution \(\mathcal C(w, D_k)\) for client \(k\) by introducing data $D_k$ in FedFMs. Assume an Oracle knows the true contribution $C_k^*$ for each client $k.$
The objective of contribution evaluation is to look for the contribution evaluation mechanishm $\mathcal C$ such that this evaluation is accurate \cite{song2019profit, wei2020efficient}, formulated as:
\begin{equation}
 \min_{\mathcal C}\ell_{c,k}(C,w_s,D_k)
 :=\sum_{k=1}^K\|\mathcal C(w_s, D_k) -C_k^*\|_2^2.
\end{equation}
Here ${\ell_{c,k}(\cdot)}$ is the loss of contribution estimation for client $k$.

\begin{figure}[htb]
        \centering
        \includegraphics[width=0.35\textwidth]{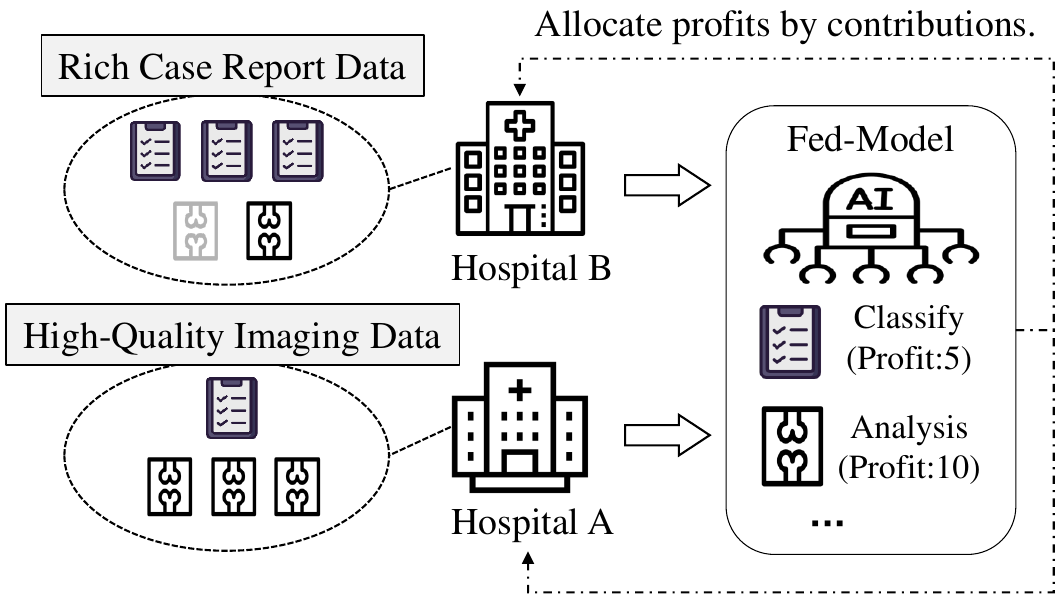} %
        \caption{Contribution Evaluation across Medical Institutions.}
        \label{fig:Q3_Contribution_Example}
        \vspace{-15pt}
    \end{figure}

\subsection{Existing Methods}
In contribution evaluation for FedFMs, the first step is to select an evaluation scheme and then chose a utility function to assess the FedFM's performance.
Prior work in contribution evaluation for FedFMs mainly focus on two aspects: \textit{(1) contribution evaluation schemes} and \textit{(2) utility functions}.

\subsubsection{Contribution Evaluation Schemes}

Existing work primarily include five schemes, \textit{i.e.,} \textit{individual}, \textit{interaction}, \textit{leave-one-out}, \textit{shapley value}, and \textit{least core}.

\textbf{Individual}.
The individual method~\cite{Individual} determines an individual's contribution in FedFMs by directly using the output of the utility function.
Formally, the individual method is defined as, $\varphi_{i}(v) = v(i)$,  where $v(\cdot)$ can be any utility function and $i$ represents the $i${\small-th} FedFMs client.
The primary limitation of this method is its inability to account for the marginal contribution.
The time complexity of the \textit{Individual} method is $\mathcal{O}(n)$.

\textbf{Leave-One-Out}. 
The leave-one-out (LOO) method~\cite{LOO1999} accounts for the marginal utility, which removes a specific individual and recalculating the evaluation value using.
It is calculated based on the difference between the overall utility of the entire FedFMs system and the utility obtained after the FedFMs client's removal.
Formally, the contribution is defined as $\varphi_{i}(v) = v(N) - v(N/\{i\})$, where $N$ denotes all the clients.

\textbf{Interaction}.
It evaluates clients' contributions in the training process, where we can vary the proportion of training data in each round to estimate the FedFMs client's contribution~\cite{RN9, RN33}.
The time complexity of \textit{Interaction} is $\mathcal{O}(n)$. 
Researchers also design several specific solutions tailored for distinct characteristics and application scenarios using interaction approach.
\citet{RN9} proposes FedCCEA.
It evaluates contributions by various FedFMs through random adjustments to the proportion of training data provided by FedFMs clients.
\citet{RN33} provided a contribution evaluation scheme using gradient-based evaluations, which ensures strong model performance and is well-suited for scenarios where the data distribution among FedFMs clients is typically statistically heterogeneous.

\textbf{Shapley-Value (SV)}. It is a classical concept in the collaborative game theory to fairly measures player's contribution~\cite{Shapley1988AVF}.
The SV provides four essential properties for contribution evaluation in FedFMs and is widely adopted as the standard evaluation scheme, such as \textit{no-free-rider, symmetric-fairness} and \textit{group-rationality}.
Though the Shapley value holds several desirable properties, it suffers from the computational complexity in $\mathcal{O}(2^n)$ and existing studies mainly adopt two kinds of approximation solutions, \textit{i.e.,} the sampling-based~\cite{RN21} and the gradient-based approximations~\cite{song2019profit, wei2020efficient}.

\textbf{Least Core}.
The core method~\cite{Sch1969nucleolus} is a classic concept widely applied in fields such as economics.
It requires contributions for any set of FedFMs clients should be no less than that of its subset.
The time complexity of \textit{Least Core} is also $\mathcal{O}(2^n)$.
Yan et al. \cite{Yan2021IfYL} used the least core and converts it as an optimization problem.
This evaluation scheme seeks to optimize the overall value distribution among all FedFM clients, prioritizing fairness at the coalition level.
   
\subsubsection{Utility Functions for Contribution Evaluation} 

The utility function measures the performance of FedFMs and prior studies can be divided into following two categories.

\textbf{Task-Specific Utility}. 
It quantifies FedFMs client's utility within specific tasks.
We review the three widely-adopted utility functions as follows.
    \textit{(i) Model Performance}~\cite{RN35}.
    It is a straightforward way to evaluate the utility of FedFMs clients by measuring their impact on model performance.
    \textit{(ii) Mutual Cross Entropy}~\cite{RN4}. It is a metric that quantifies the value of the FedFMs client's local dataset by calculating the mutual cross-entropy between the FL model's performance on local datasets and that on the test dataset.
    \textit{(iii) Influence Function}~\cite{RN12, RN13}. It is a classic technique in statistics tracing model predictions to identify the training data with most influence.

\textbf{Task-Agnostic Utility}. It provides functionality to measure the FedFMs client's contributions without being bound to specific learning tasks.
We review the commonly-used metrics as follows.
    \textit{(i) Data Size and Diversity}~\cite{RN15}. It quantifies the divergence of data distributions, with its value characterizing data value~\cite{RN16}. 
     \textit{(ii) Model Similarity}~\cite{RN18}. It assumes valuable data produces local models close to the global model and use the global model to evaluate local models.
    \textit{(iii) Mutual Information}. It takes the mutual information between local datasets 
    as the metric to evaluate the  FedFMs client's utility.

\subsection{Challenges and Potential Solutions}

\textit{Scalability and computational overhead} are significant issues in FedFMs, as existing contribution evaluation methods, such as Shapley value-based approaches, exhibit exponential time complexity. It is crucial to design efficient yet fair contribution evaluation schemes tailored for billion-scale FedFMs. Additionally, \textit{data-task contribution mapping} remains challenging in FedFMs, as traditional approaches fail to account for task-specific data value variations.

To address these challenges, potential solutions include: (1) \textit{Learned contribution evaluation}, where FedFMs are trained for contribution assessment with adjustable loss functions and penalty coefficients. (2) \textit{Crowd-sourced data contribution map} can be established, leveraging results from numerous data providers to create and refine the valuation between data and downstream tasks.

\section{
Problem 8: How to design game mechanisms in FedFMs?
}
\label{game_mechanism}

\subsection{Defining Game Mechanisms in FedFMs}
Before introducing game mechanisms in FedFMs, it is essential to clearly define the roles of the various participants. FedFMs participants are often categorized into three types: honest, semi-honest and malicious. Honest participants are the most ideal, as they strictly adhere to the established FedFMs protocols and follow the prescribed procedures for each step. Semi-honest participants, on the other hand, do not fully comply with the protocols. They might discreetly collect private data without violating the FedFMs protocol to infer other participants’ private information, but they do not actively launch attacks or collude with other participants to undermine the protocol. They pose a significant threat to the privacy of the protocol because they are not actively attacking and are difficult to detect. 
Malicious participants are easily compromised by attackers or are themselves attackers disguised as legitimate FedFMs participants, to expose sensitive private data. Semi-honest attacks are the most common. They can include gradient inversion attack \cite{zhu2019deep,geiping2020inverting}, model inversion attack \cite{he2019model}, and GAN-based attack \cite{wang2019beyond}. The most classic attacks by malicious participants include poisoning attacks \cite{tolpegin2020data} and backdoor attacks \cite{gao2020backdoor,li2022backdoor}.

\begin{figure}[htbp]
    \centering
    \includegraphics[width=0.9\linewidth]{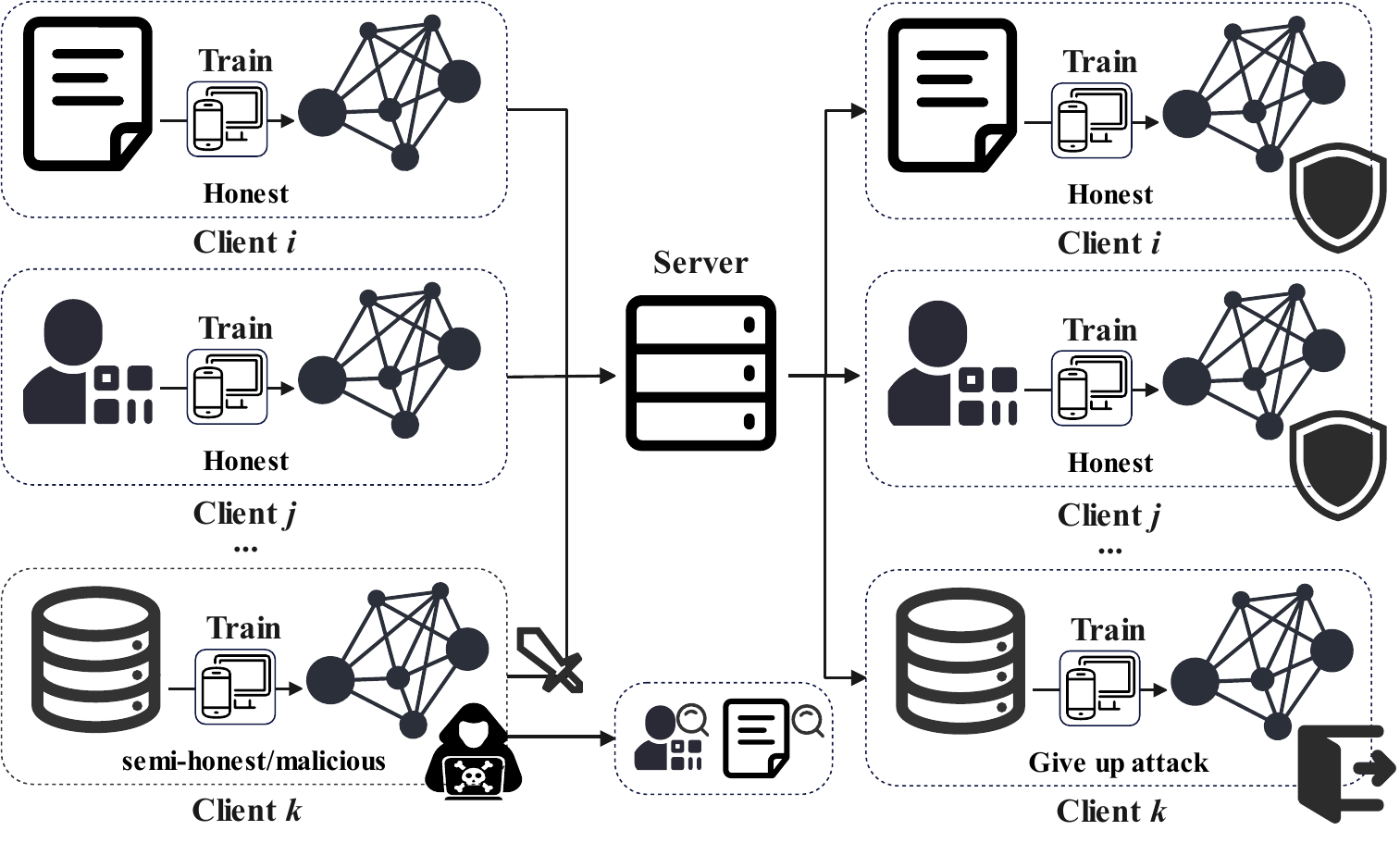}
    \captionsetup{justification=centering}
    \caption{Illustration of FedFMs Game Mechanism Design.}
    \label{fig:game_mechanism_design}
    \vspace{-5mm}
\end{figure}

A game mechanism can be designed such that the costs incurred by adversaries, including semi-honest or malicious attackers, significantly outweigh benign participants' benefits, as illustrated in Fig. \ref{fig:game_mechanism_design}. Moreover, unlike traditional FL, a game mechanism in FedFMs must balance intensive computational demands with robust privacy guarantees cross diverse participants.
These challenges manifest in three critical aspects:
\begin{itemize}
\item A fundamental tension between model performance and privacy preservation, as the requirement for extensive training data makes privacy protection substantially more complex than in smaller federated models.
\item Heightened privacy risks due to foundation models' ability to memorize and regenerate training data, rendering traditional game-theoretic approaches insufficient.
\item The need for scalable incentive mechanisms that remain economically viable while meeting substantial computational requirements.
\end{itemize}

Addressing these challenges calls for a structured game-theoretic approach in FedFMs mechanism design. Without it, the effectiveness and sustainability of FedFMs systems are significantly undermined, resulting in increased vulnerability to attacks, inefficient resource utilization and erosion of trust.

\textbf{Problem Formulation} Assume $K$ clients participate in FedFMs, and let $\mathcal{A}_k$ represent the set of actions available to participant $k$, where $a_k \in \mathcal{A}_k$ is the action chosen by participant $k$. These actions may include contributing data, reporting malicious activity, or abstaining from malicious behavior. The utility function $U_k(a_k, a_{-k})$\footnote{The parameter of Utility function $U(\cdot)$ is action $a$, which is different from the utility loss $\ell_u$ whose parameter is weight.} for participant $k$ depends on their own action $a_k$ and the collective actions of all other participants, denoted by $a_{-k}$. To capture the interplay between utility, privacy, and efficiency, the utility function is defined as follows:
\begin{equation}
\label{eq:game_utility}
U_k(a_k, a_{-k}) = r_k - c_k(a_k) + \lambda_k R_k(a_k) - \alpha P_k(a_k) - \beta E_k(a_k),
\end{equation}
where $r_k$ is the reward for honest behavior. $c_k(a_k)$ is the cost associated with the action $a_k$, which may involve computational resources or privacy risks. $R_k(a_k)$ is the reputation score with weight $\lambda_k$. $P_k(a_k)$ is the privacy loss function with weight $\alpha$. $E_k(a_k)$ is the computational efficiency cost with weight $\beta$. 

The game-theoretic equilibrium is reached when no participant can improve their utility by unilaterally changing their action, meaning the system reaches a Nash Equilibrium:
\begin{equation}
\label{eq:game_nash}
U_k(a_k^*, a_{-k}^*) \geq U_k(a_k, a_{-k}^*), \forall k \in [K], \forall a_k \in \mathcal{A}_k,
\end{equation}
where $a_k^*$ and $a_{-k}^*$ are the equilibrium strategies for participant $k$ and all other participants, respectively.

The objective of benign participants is to maximize utility and minimize cost \cite{sarikaya2019motivating,pandey2020crowdsourcing,khan2020federated}:
\begin{equation*}
\label{eq:game_strategy_1}
\begin{aligned}
&\text{benign participants:}  \\
&\max \sum_{k \text{ is benign}} U_k(a_k, a_{-k}) \quad \& \quad \min \sum_{k \text{is 
 benign}}\text{Cost}_k(a_k),
\end{aligned}
\end{equation*}
where $U_k(a_k, a_{-k})$ and $\text{Cost}_k(a_k)$ are utility and cost of benign participant $k$. Similarly, the objective of attackers is also to maximize utility and minimize cost
\begin{equation*}
\label{eq:game_strategy_2}
\begin{aligned}
&\text{Attackers:}  \\
&\min\sum_{k\text{ is 
attacker}} U_{att}(a_k, a_{-k}) \quad \& \quad \min \sum_{k\text{ is attacker}}\text{Cost}_{att}(a_k),
\end{aligned}
\end{equation*}
where $U_{att}(a_k, a_{-k})$ and $\text{Cost}_{att}(a_k)$ are utility and cost of attacker $k$. We unify the two equations as:
\begin{equation}
    \begin{aligned}
&\max \sum_{k=1}^K U_k(a_k, a_{-k}) \quad \& \quad \min \sum_{k=1}^K\text{Cost}_k(a_k),
\end{aligned}
\end{equation}

This mechanism creates a multi-layered defense by raising both technical and economic costs of attacks. Through this game theoretic framework, FedFMs establish a sustainable ecosystem where participants can safely contribute to model training while receiving fair compensation, making compliance and security the rational choice for all participants.

\subsection{Existing Methods}

Game theory, as an important socio-economic theory, has long been applied in computer science. From the perspective of game theory, \citet{zhang2024game} proposed a framework that considers defenders and attackers in FL and estimates the upper bounds of their payoffs, taking into account model utility, privacy leakage and costs. \citet{zhan2020learning} studied the motivation of edge nodes to participate in FL training. \citet{liao2024optimal} developed a mechanism that optimizes client sampling probabilities while ensuring truthful cost reporting from strategic clients. \citet{wu2020privacy} formulated privacy attacks and defenses as a privacy protection attack and defense game. They measure the defender’s payoff as privacy leakage, and the attacker’s payoff as privacy gain, thereby formulating a zero-sum game. \citet{wu2017game} transformed the trade-off between privacy and utility into a game-theoretic problem, constructing a multi-player game model to evaluate the efficiency of pure Nash Equilibria. \citet{he2024generative} explored the advantages of combining generative AI with game theory and further proposed a novel solution that combines powerful reasoning and generation capabilities of generative AI with the design and optimization of mobile networking via game theory. \citet{zhang2021more} found that differential privacy can have a favorable impact on game theory and quantify the cost of protection.

These game-theoretic efforts fundamentally aim to establish equilibrium between conflicting objectives in collaborative learning systems – protecting sensitive information while maintaining model efficacy and cost-effectiveness. To operationalize this equilibrium, researchers predominantly employ incentive mechanism design grounded in quantified payoff structures.
Based on the specific game adopted, existing methods can be divided into the following categories. 

\textbf{Stackelberg Game-based Methods}. 
Stackelberg games model FL interactions, with the server as the leader and clients as sellers setting prices. These methods optimize utility by balancing data, rewards, and resources. Prior works address fairness \cite{sarikaya2019motivating}, communication \cite{pandey2020crowdsourcing}, accuracy \cite{khan2020federated, luo2023incentive}, privacy \cite{hu2020trading}, and MEC pricing \cite{lee2020market}, though challenges like uniform pricing and unrealistic assumptions remain.

\textbf{Yardstick Competition-based Schemes}.
To reduce training delays, \citet{sarikaya2020regulating} proposed a yardstick competition where faster training earns higher rewards. While effective for synchronous FL, it oversimplifies delay factors.

\textbf{Shapley-Value (SV) Based Schemes}.
SV-based methods allocate rewards based on contribution. \citet{qu2020privacy} incentivized edge servers via gradient similarity, but ignored client input. \citet{song2019profit} introduced the Contribution Index (CI), an SV variant, but it remains limited to Horizontal FL.

\textbf{Other Incentive Mechanisms}.
Addressing payment delays, \citet{yu2020sustainable} proposed the Federated Learning Incentivizer (FLI), a real-time installment-based scheme improving fairness and revenue. Additionally, \cite{tang2023utility,tang2023competitive,tang2024bias,tang2023multi,tang2024cost,tang2024intelligent,tang2024stakeholder} modeled FL data trading as auctions, optimizing transactions based on budgets, training needs, and resources. These approaches enhance efficiency and fairness by fostering competition.

\subsection{Challenges and Potential Solutions}

\textit{Privacy cost quantification} is challenging due to complex interactions between encryption costs, potential breach losses, and diverse participant incentives, complicating the development of fair game-theoretic privacy models \cite{tang2024towards}. \textit{Information asymmetry}, limited visibility into participant payoffs and motivations, further complicates the design of incentive-compatible privacy mechanisms \cite{tang2024intelligent}, especially when balancing information sharing with security requirements. The \textit{dynamic security landscape} demands adaptive mechanisms that respond to evolving attack strategies, surpassing traditional static models. Finding the optimal balance between privacy strength, system utility, and computational efficiency remains an open challenge in privacy-preserving mechanism design.

To address these challenges, potential solutions include: (1) \textit{Adaptive game-theoretic models} leveraging reinforcement learning \cite{tang2024intelligent,zhang2021deep} can dynamically adjust strategies, improving resilience and utility. (2) \textit{Resource-aware privacy mechanisms} \cite{mishra2024resource} optimize protections based on client resources, balancing security and efficiency. (3) \textit{Incentive-compatible schemes} \cite{tang2024intelligent} use rewards and penalties to promote honest participation and deter attacks. (4) \textit{AI-driven anomaly detection} \cite{zeng2024towards} enhances security by identifying suspicious behavior in real time. Combined, these approaches strengthen FedFMs’ privacy and security in dynamic environments.

\section{Problem 9: How to Design Model Watermarking in FedFMs?}

\newcommand{\et}{\textit{et al.}}
\newcommand{\eg}{\textit{e.g.,}}
\newcommand{\ie}{\textit{i.e.,}}
\newcommand{\eq}{Eq.}
\newcommand{\fig}{Fig.}
\newcommand{\tab}{Tab.}
\newcommand{\s}{Sec.}
\newcommand{\alg}{Alg.}

\subsection{Defining Model Watermarking in FedFMs}

Model watermarking in FedFMs plays a crucial role in safeguarding intellectual property (IP), helping to ensure that FedFMs ownership is protected and preventing unauthorized use by individual clients during local training and model theft by embedding unique watermark for each client.

Watermarking in FedFMs refers to the process by which both clients and the central server collaboratively embed a watermark into a shared model to assert ownership and ensure traceability. Given that FedFMs involve multiple clients contributing to a jointly trained model while a central server orchestrates training, watermarking serves as a mechanism for both clients and the server to embed unique identifiers. This enables clients to verify their contributions and claim ownership in case of disputes, while also allowing the server to track model integrity and detect unauthorized modifications.

\textbf{Problem Formulation}. There are two categories of model watermarking:  \textit{White-Box} and \textit{Black-Box}.

\textbf{White-Box.} A binary watermark \(B \in \{0,1\}^{N_b}\) is embedded directly into the model parameters \(\theta\) (\eg~batch normalization layers) during training by adding specialized regularization terms to the loss function \cite{li2022fedipr}. During verification \(V_w\), an extractor \(E\) retrieves the watermark \(\tilde{B}\) from \(\theta\). The watermark is verified if the Hamming distance \(H(B, \tilde{B})\) is below a predefined threshold \(\epsilon_w\) as illustrated in \fig~\ref{fig: q09 watermark method}.

\textbf{Black-Box.} \(N_t\) trigger-based samples \(T = \{ (x^i_t, y^i_t) \}_{i=1}^{N_t}\) are embedded into the model \(\theta\) using a trigger objective during training \cite{li2022fedipr}. In the verification step (\(V_b\)), trigger inputs \(x_t\) are fed into the model \(\mathcal{M}\), and the outputs are compared to the corresponding labels \(y_t\) to compute the trigger accuracy \(A_t\). The watermark is confirmed if \(A_t\) exceeds a predefined threshold \(\epsilon_b\) as illustrated in \fig~\ref{fig: q09 watermark method}.

The watermark embedding process for client \( k \) minimizes a loss comprising the main task on \( \mathcal{D}_k \) and two regularization terms, \( \ell_{u,T_k}(w_s) \) and \( \ell_{u,B_k}(w_s) \), to embed trigger samples \( T_k \) and feature-based watermarks \( B_k \) in terms of the server's global model $w_s$. Given the global model $w$ at communication round \( t \), the objective formulation is defined as:

\begin{equation}
\begin{split}
      & \min_wF_{k=0,1,\cdots, K}\alpha\ell_{u,k}(w_s)+(1-\alpha)\ell_{m,k}(w_s) \\
       &:=\Big(\alpha\underbrace{\ell_{u,k}(w_s)}_{\text{utility loss}} 
    + \beta\underbrace{\ell_{u,T_k}(w_s)}_{\text{black-box}} 
    + \gamma\underbrace{\ell_{u,B_k}(w_s)}_{\text{white-box}} \Big), \\
    &s.t. \quad \ell_{e,k}(w_s)<\delta_k,
\end{split}
\end{equation}

where $\ell_{u,k}(w_s)$ and $\ell_{m,k}(w_s)$  represent the model utility and watermark loss receptively. Watermark loss aims to make the watermark embedded in the model accurately. $\ell_{u,T_k}(w_s)$ and  $\ell_{u,B_k}(w_s)$ represent black-box watermark and white-box watermark loss respectively. $F_{k=0,1,\cdots, K}$ is the aggregation mechanishm for the server and $K$ clients such as FedAvg \cite{mcmahan2017communication}.  $\alpha,\beta,\gamma$ are the respective weights reflecting the relative importance of each component and meet $\alpha+\beta+\gamma=1$.

\subsection{Existing Methods}

\begin{figure}
    \centering
    \begin{adjustbox}{max width= \linewidth}
        \begin{tabular}{c c}
             \includegraphics[width= 0.5\linewidth]{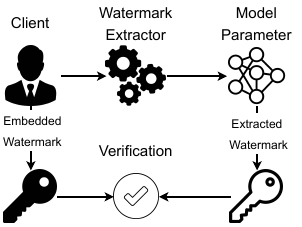}&  \includegraphics[width= 0.5\linewidth]{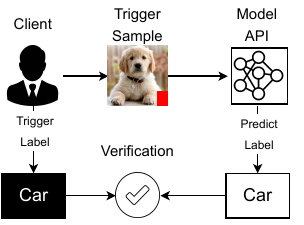}\\
             (a) White-Box& (b) Black-Box\\ 
        \end{tabular}
    \end{adjustbox}
    \caption{Illustration of watermarking methods in FedFMs.}
    \label{fig: q09 watermark method}
    \vspace{-5mm}
\end{figure}

In FedFMs, watermarking methods are categorized as \textbf{white-box} or \textbf{black-box}, as shown in \fig~\ref{fig: q09 watermark method}. White-box methods use a \textbf{feature-based} approach to embed binary strings into model parameters via regularization terms \cite{uchida2017embedding,chen2018deepmarks,darvish2019deepsigns} or passport layers \cite{fan2019rethinking, fan2021deepipr}, allowing verification through full model access. In contrast, black-box methods employ \textbf{trigger-based} verification, using adversarial backdoor training samples to verify watermarks via model outputs \cite{adi2018turning,zhang2018protecting}.

\textbf{White-Box.} FedIPR \cite{li2022fedipr} introduced a client-side algorithm embedding unique, detectable watermarks in shared models, ensuring robustness against privacy-preserving techniques while identifying freeriders to promote fairness. DUW \cite{yu2023leakedmodeltrackingip} countered client-side tampering by injecting server-side watermarks, preserving their integrity. FedSOV \cite{yang2023fedsov} enhanced resilience to ambiguity attacks by embedding client public keys as watermarks via a near-collision-resistant hash function, reinforced with digital signatures to prevent forgery. FedTracker \cite{shao2024fedtrackerfurnishingownershipverification} balanced fidelity and traceability using continual learning and a watermark discrimination mechanism, while FedCIP \cite{liang2023fedcip} improved traceability by cross-referencing multiple iterations.

\textbf{Black-Box.} WAFFLE \cite{atli2021wafflewatermarkingfederatedlearning} introduced the first black-box watermarking approach in FedFMs, relying on a trusted aggregator to pre-train the global model with a trigger set before standard FL training. FedTracker \cite{shao2024fedtrackerfurnishingownershipverification} improved WAFFLE by preventing watermarking-induced model updates from interfering with primary task training and freezing batch normalization layers to maintain distribution consistency. Client-side Backdoor \cite{9658998} allowed clients to embed trigger sets in local updates. Yang \et~ \cite{yang2023watermarkingsecurefederatedlearning} refined this by structuring multi-class trigger sets into a grid-based scheme to mitigate overfitting. PWFed \cite{electronics13214306} further enhanced robustness by dynamically generating and embedding watermarks based on varying vectors.

Note that FedTracker involves both white-box and black-box methods.

\subsection{Challenges and Potential Solutions}
\textit{Large-scale watermarking} in FedFMs faces challenges such as computational overhead on resource-limited clients, \textit{scalability issues} with massive models like LLMs and ViTs due to their size and communication costs, and the risk of watermark distortion from compression or quantization. Additionally, the \textit{diverse data modalities} handled by FedFMs (e.g., images, text, audio, video) complicate the creation of universal watermarks due to architectural and sensitivity differences. The \textit{dynamic client participation} in FedFMs further challenges the integrity and persistence of watermarks, making this an underexplored area. 

To address these challenges, potential solutions include: (1) \textit{Lightweight embedding and federated pruning} techniques \cite{jiang2022modelpruningenablesefficient, lin2022federatedpruningimprovingneural} can minimize computational demands on resource-constrained devices. (2) \textit{Resource-efficient watermarking} methods \cite{jimaging8060152}, including memory-efficient anti-pruning strategies, ensure robustness against compression techniques like quantization, pruning, and distillation. (3) \textit{Modality-agnostic watermarking }
\cite{tang2023watermarkingvisionlanguagepretrainedmodels} can embed watermarks in shared latent representations or parameter distributions, enabling compatibility across diverse data types. (4) \textit{Adaptive watermarks with self-reinforcing signatures} can enhance persistence and integrity in dynamic aggregation scenarios.

\section{Problem 10: How to to improve the efficiency in FedFMs?}

\subsection{Defining efficiency of FedFMs}
Borrowing the definitions in \cite{yao2024federated}, the efficiency of FedFMs corresponds to the three fundamental aspects of computation, communication, and storage.
Thus, the efficiency of FedFMs refers to its capability to minimize the weighted sum of these three factors, facilitating scalable and high-performance training in resource-constrained environments \cite{chen2021communication, ren2024advances}. Challenges such as high update frequencies, large model sizes, and network unreliability hinder scalability, while the resource demands of complex FMs exacerbate the strain on devices. Therefore, improving computation, communication, and storage efficiency is vital for enabling FedFMs' deployment in sectors like healthcare, finance, and IoT. 

\textbf{Problem Formulation.} This section considers three types of efficiency in FedFMs \cite{mcmahan2017communication,wang2021resource, hu2021mhat}: the computation loss, communication  loss, and model size for model $w$ at time slot $t$ as $\tilde \ell_e^t(w)$, $\hat \ell_e^t(w)$, and $\bar \ell_e^t(w)$, respectively. The objective is to minimize the overall efficiency consumption, expressed as:
\begin{equation}
\label{eq:efficiency_FM}
\begin{split}
    \min_w\sum_{t=0}^{T}&\alpha \underbrace{\tilde \ell_e^t(w)}_{\text{computation loss}}+\beta \underbrace{\hat \ell_e^t(w)}_{\text{communication loss}} + \gamma  \underbrace{\bar\ell_e^t(w)}_{\text{model size}}, \\
    &s.t. \quad \ell_{p,k}(w) <\delta_k,
    \end{split}
\end{equation}
where $T$ is training episodes; $\alpha,\beta,\gamma$ are the respective weights reflecting the relative importance of each efficiency component and meet $\alpha+\beta+\gamma=1$. This formulation ensures a balanced trade-off among the three efficiency dimensions, optimizing FedFMs for real-world, resource-constrained environments.

\subsection{Existing Methods}

\begin{figure}[t]
    \centering
     \includegraphics[width=0.4\textwidth]{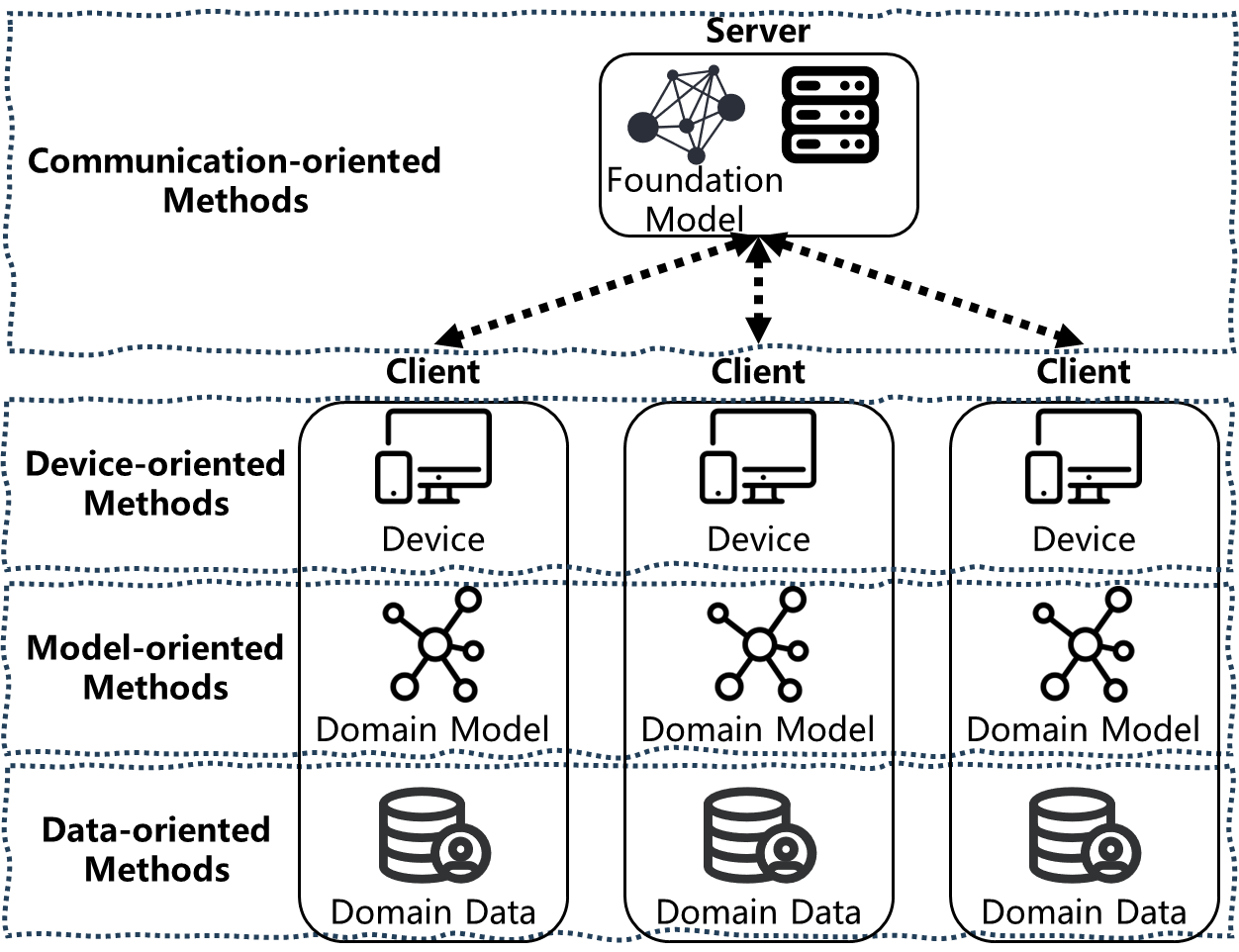}
     \caption{Summary of Efficiency in FedFMs}
     \label{fig:efficiency_fm}
     \vspace{-5mm}
 \end{figure}

A wide range of methods has been proposed to enhance the efficiency of FedFMs. Based on the key elements of FedFMs, we categorize the existing literature into a taxonomy comprising four main categories: \textit{data-oriented methods}, \textit{model-oriented methods}, \textit{device-oriented methods}, and \textit{communication-oriented methods}, as shown in Fig. \ref{fig:efficiency_fm}.

\subsubsection{Data-oriented Methods}

Data-oriented methods emphasize the importance of local data quality and structure in improving the efficiency of local FMs \cite{ye2024openfedllm, qin2024federated}. \textbf{Data selection} is essential for improving the efficiency of FedFMs pre-training and fine-tuning, which selects efficient datasets to reduce training FLOPs \cite{yao2022nlp, fawcett2024improving}. \textbf{Prompt compression} optimizes how FMs interact with prompts to generate desired outputs \cite{che2023federated}. Prompt compression accelerates FedFMs input processing by either condensing lengthy prompts or learning compact representations of prompts \cite{mu2024learning}.

\subsubsection{Model-oriented Methods}

Model-oriented methods focus on reducing the sizes and the number of arithmetic operations of FMs, i.e., model compression. \textbf{Quantization} transforms model parameters into lower-precision formats, such as fixed-point or reduced-bit representations (e.g., 8-bit or 4-bit), which can significantly reduce the data payload with minimal loss of model accuracy \cite{ren2023two, cao2024fedmq, mao2022communication}. 
\textbf{Parameter pruning} compresses FMs by removing redundant or less important model weights $W$, which can be categorized into structured pruning and unstructured pruning \cite{fan2024data,huang2024fedmef}.
\textbf{Knowledge distillation} trains a smaller “student” model to approximate the performance of a larger “teacher” model, thereby conserving computational resources and minimizing communication overhead \cite{ li2019fedmd}.
\textbf{Low-rank approximation:} Low-rank approximation compresses FMs by approximating the weight matrix with smaller low-rank matrices \cite{song2024low}.
This decomposition effectively reduces the number of parameters and enhances computational efficiency.

\subsubsection{Device-oriented Methods}

Device-oriented methods focus on optimizing the efficiency of FedFMs by tailoring strategies to the specific capabilities and constraints of client devices \cite{hou2020dynabert,laskaridis2024melting}.
\textbf{Resource-aware scheduling} enhances the efficient use of computing resources on client devices by integrating advanced scheduling policies into model training and communication processes \cite{hou2020dynabert,luo2021info,luo2022info}. 
\textbf{Framework optimization} focuses on lightweight libraries designed for local device constraints, enabling efficient deployment on devices with limited resources \cite{niu2024smartmem, kwon2023efficient}.

\subsubsection{Communication-oriented Methods}

Communication-oriented methods aim to reduce the volume of parameters uploaded from clients to the server during the aggregation phase while maintaining model performance \cite{wang2021resource, hu2021mhat}. For instance, \citet{wang2021resource} introduced Resource-efficient FL with Hierarchical Aggregation (RFL-HA) to mitigate high training time and communication overhead in FL, which organizes edge nodes into clusters to reduce data transmission. Furthermore, \citet{zheng2022aggregation} presented an FL system that enhances privacy by enabling clients to share obscured model updates via lightweight encryption. Additionally, \citet{al2023decentralized} eliminated the need for a central aggregator through a decentralized FL approach that leverages device-to-device communication and overlapped clustering, significantly reducing communication costs.

\subsection{Challenges and Potential Solutions}

FedFMs face \textit{communication bottlenecks} due to constant synchronization and model aggregation, leading to inefficiency, backhaul bottlenecks, and high latency. Additionally, there is an \textit{efficiency-performance-privacy trade-off}, as efficiency-preserving methods like HE often reduce computational efficiency, making it challenging to balance privacy and model performance. Furthermore, \textit{task redundancy} in FedFMs results in significant computational overhead from repetitive computations by multiple clients.

To address these challenges, potential solutions include: (1) \textit{Compression}: Exploring ways to minimize communication through adaptive aggregation frequencies \cite{lee2023layer}, efficient compression techniques, and advanced communication methods; (2) \textit{Hybrid Privacy-Preserving Techniques}: Developing advanced HE schemes and novel hybrid techniques to ensure efficiency and strong data privacy \cite{254465}; (3) \textit{Disentangling Tasks}: Researching methods to disentangle complex tasks and effectively assign subtasks to different clients under server coordination, such as dependent task offloading \cite{wang2021dependent}, to enhance overall efficiency.

\section{Summary}
\label{sec:summay}
This paper provides a broad and quantitative examination of ten challenging problems inherent in FedFMs, encompassing issues of foundational theory, utilization of private data, continual learning, unlearning, Non-IID and graph data, bidirectional knowledge transfer, incentive mechanism design, game mechanism design, model watermarking, and efficiency. 
In this section, we unify the objective functions of these ten challenging problems in FedFMs into the following equation: 
\begin{equation}\label{eq:unifed}
    \begin{split}
        \min&_{w_g=(w_s, \{w_k\}_{k=1}^K), \{a_k\}_{k=1}^K,F,C}F_{k\in[K], t\in [T] }\Big( \\
        &\alpha_1\underbrace{\ell_{u,k}(w_s,w_k,a_k,C,\{D_k^t\}_{t=1}^T, G)}_{\text{Utility loss}}, \\
        &\alpha_2\underbrace{\ell_{e,k}(w_s,w_k,a_k,C,\{D_k^t\}_{t=1}^T,G)}_{\text{Efficiency loss}}\\
               &\alpha_3\underbrace{\ell_{m,k}(w_s,w_k,a_k,C,\{D_k^t\}_{t=1}^T, G)}_{\text{Watermark loss}}, \\
              &\alpha_4\underbrace{\ell_{c,k}(w_s,w_k,a_k,C,\{D_k^t\}_{t=1}^T, G) }_{\text{Contribution loss}}, \\
              & \alpha_5\underbrace{\ell_{p,k}(w_s,w_k,a_k,C,\{D_k^t\}_{t=1}^T, G) \Big)}_{\text{Privacy loss}} \\
            & s.t.\quad  \underbrace{\ell_u(w_g) + \ell_p(w_g)+\ell_e(w_g)>0}_{\text{No free lunch constraint}},
    \end{split}
\end{equation}
where $\ell_{u,k}$, $\ell_{e,k}$, $\ell_{m,k}$, $\ell_{c,k}$, and $\ell_{p,k}$ denote the utility loss, efficiency loss, watermark loss, contribution loss, and privacy loss, respectively. These losses are defined in terms of several parameters: aggregation mechanism $F$, contribution evaluation mechanism $C$, server's model $w_s$, client $k$'s model $w_k$, and action $a_k$. The coefficients $\alpha_1$, $\alpha_2$, $\alpha_3$, $\alpha_4$ and $\alpha_5$ take values in the range $[0, 1]$ and meet $\sum_{i=1}^5\alpha_i=1$. 
Specifically, when $\alpha_1 = 1$ and $\alpha_2=\alpha_3=\alpha_4 = \alpha_5= 0$, Eq.  \eqref{eq:unifed} minimizes the utility loss through the utilization of private data, continual learning data, graph data (i.e., problem 2, 3, 5, 6, 8) When $\alpha_2 = 1$ and $\alpha_1=\alpha_3=\alpha_4 = \alpha_5= 0$, Eq.  \eqref{eq:unifed} optimizes the efficiency loss (i,e., problem 10). When $\alpha_3 = 1$ and $\alpha_1=\alpha_2=\alpha_4 = \alpha_5= 0$, Eq.  \eqref{eq:unifed} optimizes the watermark loss (i.e., problem 9). When $\alpha_4 = 1$ and $\alpha_1=\alpha_2=\alpha_3 = \alpha_5= 0$, Eq.  \eqref{eq:unifed} optimizes the contribution loss (i.e., problem 7). When $\alpha_5 = 1$ and $\alpha_1=\alpha_2=\alpha_3 = \alpha_4= 0$, Eq.  \eqref{eq:unifed} optimizes the privacy loss (i.e., problem 4). Finally, the constraint of Eq.  \eqref{eq:unifed} is based on the no free lunch theory of FedFMs (i.e., problem 1). To the best of our knowledge, \textit{this unified equation is the first to present the key problems of FedFMs from a mathematical perspective}.

In summary,  this paper delves into the ten challenging problems faced by FedFMs, which are outlined in Table \ref{tab:framework}. 
Each of the ten problems in FedFMs can be formulated as an optimization problem with a specific objective function. These objective functions, when unified, enable a holistic analysis of the trade-offs and boundaries across different dimensions. The distributed learning paradigm in FedFMs, which emphasizes mutual learning between large FMs and smaller DMs, exhibits high generalization capabilities. The optimization objective in this paradigm is to achieve a balance across multiple sub-objectives.
By overcoming these challenges, this paper underscores the importance of advancing FedFMs to unlock their full potential.
Our ultimate goal is to foster a distributed model ecosystem that emphasizes mutual learning between FMs in the server and local DMs in the clients, ensuring robust, efficient, and privacy-preserving FedFMs.  This will not only advance the theoretical foundations of FedFMs but also facilitate their widespread adoption in real-world applications, paving the way for revolutionary advancements across various domains, including healthcare, finance, and IoTs.

\appendices

\ifCLASSOPTIONcaptionsoff
  \newpage
\fi

\footnotesize
\bibliographystyle{IEEEtranN}
\bibliography{reference_new1}

\vfill

\end{document}